\documentclass[letterpaper]{article} % DO NOT CHANGE THIS
\usepackage{aaai2026}  % DO NOT CHANGE THIS
\usepackage{times}  % DO NOT CHANGE THIS
\usepackage{helvet}  % DO NOT CHANGE THIS
\usepackage{courier}  % DO NOT CHANGE THIS
\usepackage[hyphens]{url}  % DO NOT CHANGE THIS
\usepackage{graphicx} % DO NOT CHANGE THIS
\usepackage{natbib}  % DO NOT CHANGE THIS AND DO NOT ADD ANY OPTIONS TO IT
\usepackage{caption} % DO NOT CHANGE THIS AND DO NOT ADD ANY OPTIONS TO IT
\usepackage{algorithm}
\usepackage{algorithmic}

\usepackage{newfloat}
\usepackage{listings}
\DeclareCaptionStyle{ruled}{labelfont=normalfont,labelsep=colon,strut=off} % DO NOT CHANGE THIS
\floatstyle{ruled}
\newfloat{listing}{tb}{lst}{}
\floatname{listing}{Listing}

\usepackage[utf8]{inputenc} % allow utf-8 input
\usepackage[T1]{fontenc}    % use 8-bit T1 fonts
\usepackage{url}            % simple URL typesetting
\usepackage{booktabs}       % professional-quality tables
\usepackage{amsfonts}       % blackboard math symbols
\usepackage{nicefrac}       % compact symbols for 1/2, etc.
\usepackage{microtype}      % microtypography
\usepackage{xcolor}         % colors
\usepackage{amsmath}
\usepackage{amssymb}
\usepackage{amsfonts}
\usepackage{caption}
\usepackage{color}
\usepackage{booktabs}
\usepackage{multirow}
\usepackage{subfigure}
\usepackage{pifont}
\usepackage{subfigure}
\usepackage{graphicx}
\usepackage{pifont}
\usepackage{colortbl}
\usepackage{caption}
\usepackage{subcaption}
\usepackage{graphicx}
\usepackage{enumitem}          
\usepackage[capitalize,noabbrev]{cleveref}
\definecolor{LightCyan}{rgb}{0.9, 0.9, 0.9}
\usepackage{soul}
\usepackage{multirow}
\usepackage{rotating}
\usepackage{longtable}

\title{Learning Cell-Aware Hierarchical Multi-Modal Representations for Robust Molecular Modeling}
\author {
    Mengran Li\textsuperscript{\rm 1,2,3,4}\thanks{These authors contributed equally.},
    Zelin Zang\textsuperscript{\rm 2,3,4}$^*\!$,
    Wenbin Xing\textsuperscript{\rm 1},\\
    Junzhou Chen\textsuperscript{\rm 1}\thanks{Corresponding author.},
    Ronghui Zhang\textsuperscript{\rm 1},
    Jiebo Luo\textsuperscript{\rm 3},
    Stan Z. Li\textsuperscript{\rm 2},
}
\affiliations {
    \textsuperscript{\rm 1}School of Intelligent Systems Engineering, Sun Yat-sen University\\
    \textsuperscript{\rm 2}Westlake University\\
    \textsuperscript{\rm 3}Center for Artificial Intelligence and Robotics (CAIR), Hong Kong Institute of Science and Innovation (HKISI)\\
    \textsuperscript{\rm 4}Center for Integrated Circuits and Artificial Intelligence, Tsientang Institute for Advanced Study\\
    limr39@mail2.sysu.edu.cn, zangzelin@westlake.edu.cn, chenjunzhou@mail.sysu.edu.cn\\
}

\usepackage{bibentry}
\begin{document}

\maketitle

\begin{abstract}
Understanding how chemical perturbations propagate through biological systems is essential for robust molecular property prediction. While most existing methods focus on chemical structures alone, recent advances highlight the crucial role of cellular responses such as morphology and gene expression in shaping drug effects. However, current cell-aware approaches face two key limitations: (1) modality incompleteness in external biological data, and (2) insufficient modeling of hierarchical dependencies across molecular, cellular, and genomic levels. We propose \textbf{CHMR} (Cell-aware Hierarchical Multi-modal Representations), a robust framework that jointly models local-global dependencies between molecules and cellular responses and captures latent biological hierarchies via a novel tree-structured vector quantization module. Evaluated on nine public benchmarks spanning 728 tasks, CHMR outperforms state-of-the-art baselines, yielding average improvements of \textbf{3.6\%} on classification and \textbf{17.2\%} on regression tasks. These results demonstrate the advantage of hierarchy-aware, multimodal learning for reliable and biologically grounded molecular representations, offering a generalizable framework for integrative biomedical modeling. 
The code is in \url{https://github.com/limengran98/CHMR}.
\end{abstract}

\section{Introduction}\label{sec:I}

\begin{figure}[ht]
    \centering
    \includegraphics[width=0.48\textwidth]{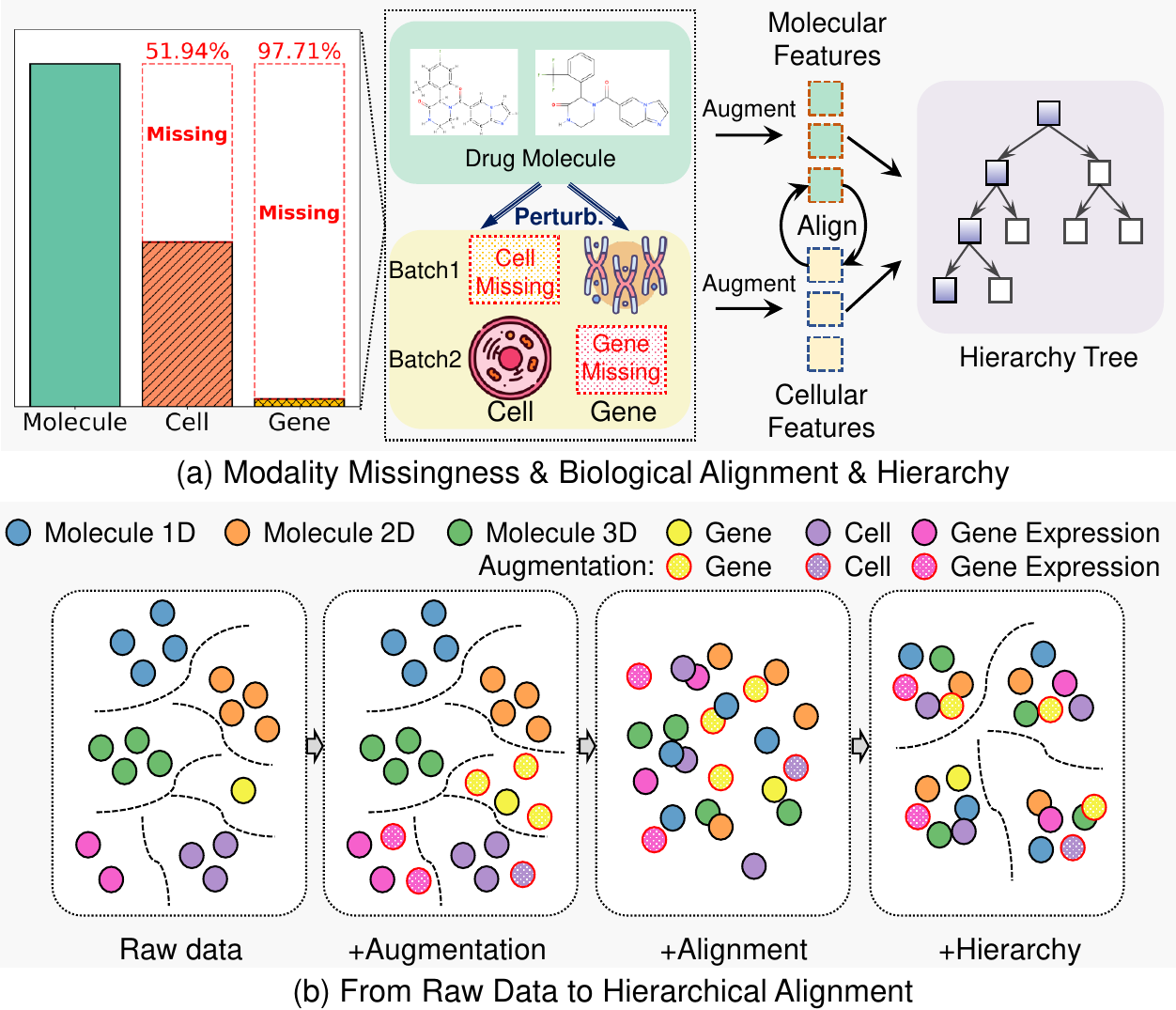}
    \caption{The motivation of this paper. Molecular perturbations trigger cellular or genetic changes, but modality incompleteness is common. Through augmentation, alignment, and hierarchical modeling, multi-modal representations are progressively organized and structured.}
    \label{fig_intro}
\end{figure}

Predicting molecular properties, such as activity \cite{liu2023using}, toxicity \cite{deng2023systematic}, and side effects \cite{zhang2025artificial}, plays a crucial role in accelerating drug development and ensuring the safety of drug candidates. With the rapid development of deep learning technologies, the prediction of drug molecular properties has increasingly become automated \cite{liu2022pre, zhou2023unimol}, significantly reducing both development time and resource consumption \cite{chen2025pretraining}. Considering the inherent graph structure of drug molecules, predicting using internal molecular features (such as atomic properties, interatomic bonds, and three-dimensional structures) has become feasible. For example, some methods pre-train on large, unlabeled molecular datasets using perturbation or augmentation strategies \cite{jiang2025adaptive, hu2025mol, jia2025association}, enabling the transfer of learned knowledge to various downstream tasks.

While structure-based models have proven effective, they often fail to capture biological responses, including gene expression and cell morphology, that are induced by molecular interactions. To complement this, researchers have started incorporating biological context into molecular representation learning \cite{wang2023removing, liu2025learning}. When molecules bind to cellular targets, they trigger signaling cascades that can alter gene expression \cite{himmelstein2017systematic, chandrasekaran2023jump} and affect cell morphology \cite{bray2016cell, bray2017dataset}. By learning cross-modal alignment and building a shared representation space, the differences between biological modalities are bridged, thereby improving the semantic understanding of molecules. Consequently, incorporating cell-aware molecular representation learning is essential for understanding the mechanisms by which molecules exert their effects and for designing safer and more effective compounds \cite{moshkov2023predicting}.

Despite recent advances, two critical challenges remain (Figure~\ref{fig_intro}(a)). \textbf{First, biological modality missingness is pervasive and asymmetric.} While molecular structure data are typically complete, associated cellular phenotypes or gene expression profiles are frequently unavailable due to experimental limitations or cost constraints~\cite{gatto2023initial, zhang2023distilling}. For instance, a compound may have morphological cell readouts but lack transcriptomic data, or vice versa. This leads to distributional shifts and modality imbalance, which undermine model robustness and limit generalization in practical scenarios. 

\textbf{Second, biological modalities exhibit hierarchical dependencies that are difficult to capture.} Molecular perturbations initiate cascades of interactions across biological layers—from chemical structures to cellular processes to gene expression programs. Yet many existing models treat these modalities in a flattened latent space and focus on instance-level alignment~\cite{sanchez2023cloome, gulati2025profiling}, failing to capture multi-hop semantic relationships and cross-layer dependencies essential for modeling cross-scale biological mechanisms.

To address these challenges, we propose \textbf{CHMR} (Cell-aware Hierarchical Multi-modal Representations), a unified framework for robust and interpretable molecular representation learning across multiple biological modalities. CHMR is designed to address two key limitations in existing cell-aware methods. First, to handle pervasive and asymmetric missingness in biological modalities, CHMR captures both local and global dependencies between molecules and their cellular responses, and enforces semantic consistency between molecular features and available cell- and gene-level information. Second, to model cross-scale biological mechanisms, CHMR incorporates a tree-structured vector quantization module that encodes latent hierarchies across molecules, cells, and genes. This design preserves biologically meaningful structures, mitigates information flattening, and enhances interpretability and generalization. Figure~\ref{fig_intro}(b) illustrates how hierarchical modeling helps align and organize multi-scale biological signals across modalities. Our key contributions are as follows:
\begin{itemize}
    \item We propose a unified framework that jointly models molecular structures, cellular phenotypes, and gene expression profiles, enabling robust and generalizable representation learning under missing biological modalities.

    \item We introduce a tree-structured vector quantization module to capture hierarchical dependencies among molecules, cells, and genes, facilitating biologically grounded fusion and improved cross-modal interpretability.

    \item We validate CHMR on 728 molecular property prediction tasks across nine benchmark datasets. CHMR consistently outperforms state-of-the-art baselines, achieving average improvements of \textbf{3.6\%} on classification and \textbf{17.2\%} on regression tasks.
\end{itemize}

\paragraph{Related Work}  We compare CHMR with structure models and cell-responsive models; \textit{see Appendix A.}

\section{Preliminaries and Notation}\label{sec:III}
\paragraph{Definition 1: Cell-aware molecular representation learning} 
Let $\mathcal{V} = \{v_i\}_{i=1}^N$ denote the set of $N$ molecules. For each molecule $v_i \in \mathcal{V}$, we define its multi-modal feature set as $\mathbf{X}_i = \{ \mathbf{x}_i^{\xi} \mid \xi \in \mathcal{M} \cup \mathcal{C} \}$, where $\xi$ indexes a generalized modality. The set $\mathcal{M}$ denotes molecular modalities, each $m \in \mathcal{M}$ corresponding to structural molecular features $\mathbf{x}_i^m$ such as molecular fingerprints encoding substructure information~\cite{rogers2010extended}, graph-based representations derived from molecular topology via graph neural networks~\cite{wang2022molecular}, or geometric features obtained from molecular conformations~\cite{zhou2023unimol}. The set $\mathcal{C}$ denotes external cellular modalities, each $c \in \mathcal{C}$ representing a molecular-induced biological response $\mathbf{x}_i^c$, such as cell morphology profiles from imaging~\cite{bray2016cell}, gene perturbation features~\cite{chandrasekaran2023jump}, or gene expression data~\cite{himmelstein2017systematic}.

\paragraph{Definition 2: Missing external cellular modality} 
The issue of missing cellular modalities arises due to experimental constraints or cost limitations. Formally, let $\mathcal{C}^{\text{obs}} \subseteq \mathcal{C}$ represent the set of observable external cellular modalities, and let $\mathcal{C}^{\text{miss}} = \mathcal{C} \setminus \mathcal{C}^{\text{obs}}$ denote the missing modalities. In this study, we assume that all molecular modalities $\{ \mathbf{x}_i^m \mid m \in \mathcal{M} \}$ are fully available, with the focus being on the scenario where external cellular modality features $\{ \mathbf{x}_i^c \mid c \in \mathcal{C}^{\text{miss}} \}$ are missing. Different molecules may exhibit different patterns of missing cellular modalities. For example, some molecules may lack cellular phenotype features, while others may lack gene expression data (Figure \ref{fig_intro}). Missing cellular modality features are replaced with placeholders, like zero vectors or global averages, to ensure consistent feature matrix dimensions.

\paragraph{Research Objectives} 
Our objective is to learn robust molecular representations that support semantic completion, preserve molecule–cell–gene hierarchical consistency under missing modalities, and generalize to downstream property prediction tasks such as solubility, activity, and toxicity.

\begin{figure*}[th]
    \centering
    \includegraphics[width=1\textwidth]{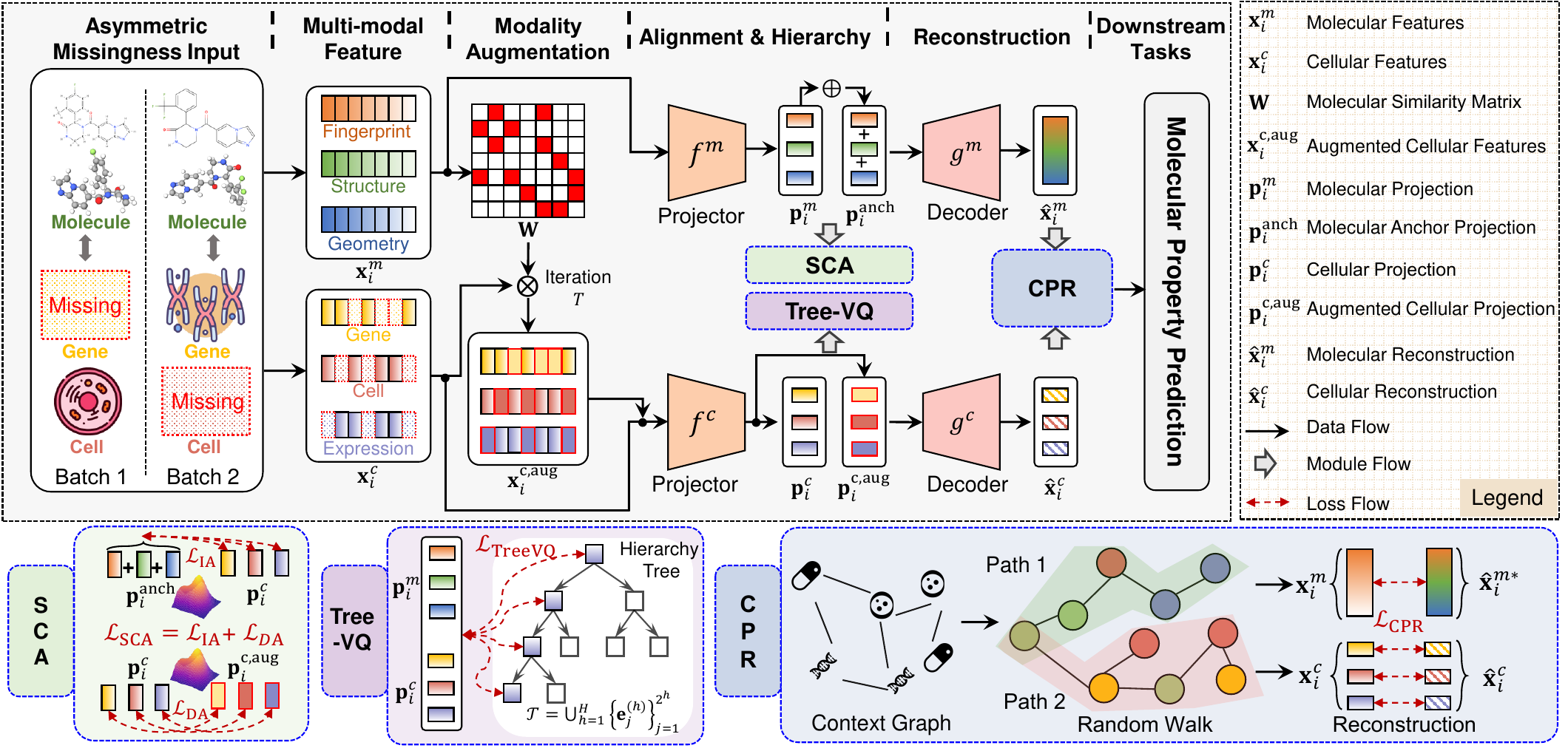}
    \caption{Overview of the CHMR framework for robust molecular property prediction under missing biological modalities. CHMR performs modality augmentation via structure-aware propagation, followed by (1) semantic consistency alignment (SCA) to align molecular and cellular modalities, (2) tree-structured vector quantization (Tree-VQ) to capture hierarchical biological semantics, and (3) context propagation reconstruction (CPR) to enhance generalization through cross-modal context.}
    \label{fig_framework}
\end{figure*}

\section{Methodology}\label{sec:IV}

We propose the CHMR framework to handle multi-modal missing data and capture hierarchical dependencies between molecules and cells, ensuring accurate prediction of molecular properties. The architecture is shown in Figure \ref{fig_framework}.

\subsection{Modality Augmentation}

To address incomplete and asymmetric biological modalities (Figure~\ref{fig_intro}(a)), we propose a structure-aware propagation strategy for modality augmentation that captures both local and global molecular context, mitigating the limitations of mean or $K$-NN imputation.

Given a molecule set $\mathcal{V}=\{v_i\}_{i=1}^N$, we construct a pairwise similarity matrix $\mathbf{W} \in \mathbb{R}^{|\mathcal{V}| \times |\mathcal{V}|}$, where $\mathbf{W}_{ij}$ denotes the structural similarity between molecules $v_i$ and $v_j$. For each molecule $v_i$, only its top-$K$ nearest neighbors $\mathcal{N}_K(v_i)$ are retained to enhance locality.

For each external modality $c \in \mathcal{C}$ of molecule $v_i$, we perform iterative propagation inspired by Dirichlet energy minimization~\cite{rossi2022unreasonable}:
\begin{equation}
\mathbf{x}^{c,(T)}_{i} =
\begin{cases}
\mathbf{x}^{c}_{i}, & c \in \mathcal{C}_i^{\mathrm{obs}}, \\[1ex]
\sum\limits_{j \in \mathcal{N}_K(v_i)} \mathbf{W}_{ij}\, \mathbf{x}^{c,(T-1)}_{j}, & c \in \mathcal{C}_i^{\mathrm{miss}}.
\end{cases}
\end{equation}

Observed modalities retain their original features, while missing ones are iteratively estimated via neighbor aggregation. The final augmented feature $\mathbf{x}_i^{c, \text{aug}}$ incorporates local structural context.

\subsection{Semantic Consistency Alignment}

To mitigate semantic discrepancies between molecular and external cellular modalities, especially those amplified by modality augmentation, we introduce the SCA module.

We begin by projecting each modality into a shared latent space:
\begin{equation}
\mathbf{p}_i^{m} = f^{m}(\mathbf{x}_i^{m}), \quad
\mathbf{p}_i^{c} = f^{c}(\mathbf{x}_i^{c}), \quad
\mathbf{p}_i^{c, \text{aug}} = f^{c}(\mathbf{x}_i^{c, \text{aug}}),
\end{equation}
where $f^{m}(\cdot)$ and $f^{c}(\cdot)$ denote the projectors for molecular and cellular modalities, respectively, mapping features into a common representation space.

\paragraph{Sample-level Alignment}
We enforce alignment between molecular and cellular representations using an InfoNCE-style contrastive loss. Molecular features are aggregated into an anchor vector $\mathbf{p}_i^{\text{anch}} = \sum_{m \in \mathcal{M}} \mathbf{p}_i^{m}$, and paired with $\mathbf{p}_i^{c}$ as positives:
\begin{equation}
\mathcal{L}_{\mathrm{IA}} = -\frac{1}{|\mathcal{V}|} \sum_{i \in \mathcal{V}} \log \frac{\exp\left(\cos (\mathbf{p}_i^{\text{anch}}, \mathbf{p}_i^{c} ) / \tau\right)}{\sum_{j \in \mathcal{V}} \exp\left(\cos ( \mathbf{p}_i^{\text{anch}}, \mathbf{p}_j^{c} ) / \tau\right)},
\end{equation}
where $\tau > 0$ is the temperature parameter.

\paragraph{Distribution-level Alignment}
To counteract potential distributional shifts introduced by propagation-based augmentation, we adopt a VICReg-style loss to align augmented and original cellular features:
\begin{equation}
\mathcal{L}_{\mathrm{DA}} = \frac{1}{|\mathcal{V}|} \sum_{i \in \mathcal{V}} \mathrm{VICReg}\left(\mathbf{p}_i^{c}, \mathbf{p}_i^{c, \text{aug}}\right),
\end{equation}
where $\mathrm{VICReg}(\cdot)$~\cite{bardes2022vicreg} enforces semantic and statistical consistency via invariance, variance, and covariance regularization. (\textit{See Appendix B.2 for details.})

The overall SCA objective combines both levels of alignment: $\mathcal{L}_{\mathrm{SCA}} = \mathcal{L}_{\mathrm{IA}} + \mathcal{L}_{\mathrm{DA}}$.

\subsection{Tree-Structured Vector Quantization}

Biological responses across molecular, cellular, and genomic levels are inherently hierarchical~\cite{gulati2025profiling}. To capture cross-scale semantics between molecular structures and biological signals, we introduce the Tree-VQ module for hierarchical representation learning. Unlike flat alignment, Tree-VQ interprets the layers of the tree as representing biological levels ranging from molecular fingerprints (shallow) to cellular phenotypes and gene expression (deep), aligned with the hierarchical responses illustrated in Figure~\ref{fig_intro}(b).

\paragraph{Multimodal Collaborative Tree Structure}
Let $\mathbf{p}^\xi \in \mathbb{R}^{d_\xi}$ denote the projected feature vector of modality $\xi$, where $\xi \in \mathcal{M} \cup \mathcal{C}$. Tree-VQ constructs a binary tree $\mathcal{T}=\bigcup_{h=1}^H\mathcal{E}^{h}$ of depth $H$, where $\mathcal{E}^{h}=\left\{\mathbf{e}_j^{h}\right\}_{j=1}^{2^h}$ represents the set of embeddings at level $h$, and $\mathbf{e}_j^{h}$ denotes the $j$-th node embedding at level $h$. Crucially, this tree is shared across all modalities, enabling heterogeneous features to be jointly embedded and routed through a unified semantic hierarchy. 

At each level $h$, we compute the cosine distance between $\mathbf{p}^{\xi}$ and candidate tree nodes:
\begin{equation}
\delta_{j}^{(\xi,h)} = 1-\cos\left(\mathbf{p}^{\xi}, \mathbf{e}_j^{h}\right).
\end{equation}

To ensure that the quantization path follows the tree hierarchy, a routing mask is applied. If the parent node selected at level $h-1$ has index $j_{\mathrm{par}}^{\xi,h-1}$, only its two child nodes $\{2j_{\mathrm{par}}^{\xi,h-1},\, 2j_{\mathrm{par}}^{\xi,h-1}+1\}$ are considered:
\begin{equation}
\tilde{\delta}_{j}^{\xi,h} =
\begin{cases}
\delta_{j}^{\xi,h}, & j \in \{2j_{\mathrm{par}}^{\xi,h-1}, 2j_{\mathrm{par}}^{\xi,h-1}+1\},\\[0.8ex]
+\infty, & \text{otherwise}.
\end{cases}
\end{equation}

\paragraph{Hierarchical Quantization and Symmetric VQ Loss}
Tree-VQ performs hierarchical vector quantization by assigning each projected feature vector to the closest tree node:
\begin{equation}
j^{*\xi,h}=\arg\min_j \tilde{\delta}_j^{\xi,h}, \quad
\mathbf{q}^{\xi,h}=\mathbf{e}_{j^{*\xi,h}}^{h}.
\end{equation}

To enforce bidirectional consistency between the encoder and tree nodes, we define a symmetric VQ loss:
\begin{equation}
\begin{aligned}
\mathcal{A}\left(\mathbf{p}^{\xi}, \mathbf{q}^{\xi,h}\right)
&=1-\cos\left(\mathrm{sg}\left[\mathbf{q}^{\xi,h}\right], \mathbf{p}^{\xi}\right)\\
&+\eta\left(1-\cos\left(\mathbf{q}^{\xi,h}, \mathrm{sg}\left[\mathbf{p}^{\xi}\right]\right)\right),
\end{aligned}
\end{equation}
where $\mathrm{sg}[\cdot]$ denotes the stop-gradient operation to ensure differentiability, and $\eta$ controls the reverse commitment weight.

Finally, the Tree-VQ loss is defined by averaging across modalities and tree levels:
\begin{equation}
\mathcal{L}_{\mathrm{TreeVQ}}=\frac{1}{|\{\xi\}|}\sum_{\xi}\frac{1}{H}\sum_{h=1}^{H}\mathcal{A}\left(\mathbf{p}^{\xi}, \mathbf{q}^{\xi,h}\right).
\end{equation}

\subsection{Context-Propagation Reconstruction}

To compensate for the lack of explicit supervision in modality augmentation, the CPR module introduces structure-aware reconstruction guided by cross-modal contextual signals. Following~\cite{liu2025learning}, we utilize a contextual graph \(\mathcal{H} = (\mathcal{U}, \mathcal{R})\), where \(\mathcal{U}\) includes molecules and external biological modalities, and each relation \((u_i, u_j) \in \mathcal{R}\) encodes a potential association. Importantly, these relations incorporate biological priors, such as known molecular perturbation–response pairs, functional associations, and shared regulatory pathways~\cite{himmelstein2017systematic, chandrasekaran2023jump}. Each relation is assigned a normalized weight \(\beta \in [0,1]\), indicating semantic strength derived from experimental, structural, or statistical sources.

On graph $\mathcal{H}$, we perform a random walk of length $L$ starting from each node $u_i$, yielding a path $(u_i,u_{i_1},...,u_{i_L})$. The propagation weights $\beta_{i,l}$ are accumulated along the path to quantify the influence of neighboring nodes.

We then use the features projected into the shared latent space as inputs to decoders for reconstructing both molecular and external modalities:
\begin{equation}
\hat{\mathbf{x}}_i^{\mathrm{anchor}} = g^{m}\big(\mathbf{p}_i^{\text{anch}}\big), \quad
\hat{\mathbf{x}}_i^{c} = g^{c}\big(\mathbf{p}_i^{c}\big),
\end{equation}
where $g^{m}(\cdot)$ and $g^{c}(\cdot)$ are the decoders for molecular and biological modalities, respectively.

To enhance reconstruction fidelity under cross-modal and hierarchical contexts, we define a unified reconstruction loss that adapts to feature types:
\begin{equation}
\mathcal{L}_{\mathrm{CPR}} = 
-\frac{1}{|\mathcal{V}|} 
\sum_{i\in\mathcal{V}} \sum_{l=0}^{L} 
\beta_{i,l}\,
\mathcal{D}\!\left(\hat{\mathbf{x}}_{u_{i_l}}, \mathbf{x}_{u_{i_l}}\right),
\end{equation}
where $\mathcal{D}(\cdot,\cdot)$ denotes the reconstruction discrepancy, using binary cross-entropy (BCE) for discrete features and mean-squared error (MSE) for continuous ones.

\subsection{Overall Objective}

\paragraph{Pretraining} To leverage the synergy of all components, we jointly optimize the loss functions of each module as:
\begin{equation}
\mathcal{L}_{\mathrm{total}} = \mathcal{L}_{\mathrm{CPR}} + \lambda_1 \mathcal{L}_{\mathrm{SCA}}  + \lambda_2 \mathcal{L}_{\mathrm{TreeVQ}},
\end{equation}
where $\lambda_1$ and $\lambda_2$ are balancing hyperparameters that control the relative importance of each module's contribution. \textit{See Appendix B.4 for time and space complexity analysis.}

\paragraph{Downstream Evaluation} 
For molecular property prediction, we freeze the pre-trained CHMR backbone and train a lightweight, task-specific prediction head $g_\theta(\cdot)$ to perform supervised prediction. \textit{See Appendix~C for details.}

\begin{table*}[htpb]\small
  \centering
  \renewcommand{\arraystretch}{0.9}
  \renewcommand{\tabcolsep}{0.3mm}
  \begin{tabular}{c|cc|ccc|ccccccc}
    \toprule
    ~ &
    \multicolumn{2}{c|}{\textbf{Dataset}} & 
    \textbf{ChEMBL} & 
    \textbf{ToxCast} & 
    \textbf{Broad}& 
    \multicolumn{7}{c}{\textbf{Biogen~(MAE $\times 100 \downarrow$ )}} \\
     &\multicolumn{2}{c|}{(\textbf{Molecule} / \textbf{Task})} & 
    (\textbf{2355/41}) & 
    (\textbf{8576/617}) & 
    (\textbf{6567/32})& 
    \multicolumn{7}{c}{(\textbf{3521 / 6})}\\
    \midrule
    \textbf{}&\textbf{Method} & \textbf{Venue} & \textbf{AUC\% $\uparrow$} & \textbf{AUC\% $\uparrow$} & \textbf{AUC\% $\uparrow$} & \textbf{Avg. $\downarrow$ } & \textbf{HLM $\downarrow$}  & \textbf{RLM $\downarrow$}  & \textbf{ER $\downarrow$} & \textbf{Solubility $\downarrow$} & \textbf{hPPB $\downarrow$} & \textbf{rPPB $\downarrow$}\\
    \midrule
    % ----- Single Modal -----
    \multirow{13}{*}{\rotatebox{90}{\textbf{Single-Modal}}} 
    &MLP  & JCIM'10 & 76.8±2.2 & 57.6±1.0 & 63.3±0.3 & 66.2±2.4 & 66.1±2.6 & 69.5±3.0 & 56.8±2.3 & 56.5±4.2 & 74.6±6.2 & 73.7±7.3  \\
    &RF   & JCIM'10 & 54.7±0.7 & 52.3±0.1 & 55.5±0.1& 52.8±0.2 & 44.2±0.1 & 51.6±0.1 &44.2±0.1 & 42.0±0.2 & 67.7±0.7 & 66.9±0.9   \\
    &GP  & JCIM'10 & 51.0±0.0 & OOM & 50.6±0.0& 60.0±0.0 & 51.3±0.0 & 61.6±0.0 & 59.5±0.0 & 49.7±0.0 & 68.8±0.0 & 69.3±0.0 \\
    &AttrMask & ICLR'20 & 73.9±0.5 & 63.1±0.8 & 59.8±0.2 & 67.3±0.3 & 82.4±1.1& 99.1±1.2 & 49.8±0.7 & 51.7±1.0 & \underline{57.9±0.6} & 62.6±0.5  \\
    &ContextPred & ICLR'20 & 77.0±0.5 & 63.0±0.6 & 60.0±0.2& 68.5±0.9 & 85.0±7.9 & 96.5±3.7& 49.7±0.4 & 55.1±2.7 & 61.4±1.8 & 63.1±0.5  \\
    &EdgePred & ICLR'20 & 75.6±0.5 & 63.5±1.1 & 59.9±0.2 & 67.8±0.9 & 81.2±10.2  & 99.1±6.9& 48.0±0.5 & 53.5±2.8 & 62.2±1.8 & 62.9±0.7\\
    &GraphCL & NeurIPS'20 & 75.6±1.6 & 52.2±0.2 & 67.2±0.5 & 53.9±0.6 & 43.8±0.3& 49.6±0.3 & 45.4±0.6 &40.6±0.5 & 76.7±1.0 & 67.1±2.2 \\
    &GROVER & NeurIPS'20 & 73.3±1.4 & 53.1±0.4 & 66.2±0.1 & 54.9±1.6 & 44.5±0.4 & 52.6±0.3 & 46.5±0.7 & 41.7±0.6 & 73.2±5.7 & 71.0±4.3\\
    &JOAO & ICML'21 & 75.1±1.0 & 52.3±0.2 & 67.3±0.4 & 55.0±0.8 & 44.5±0.5 & 51.4±0.6& 47.6±0.5 & 40.6±0.2 & 74.3±2.8 & 71.5±2.6  \\
    &MGSSL & NeurIPS'21 & 75.1±1.1 & 64.2±0.2 & 66.9±0.5 & 53.2±0.3 & 44.8±0.6&41.5±0.2 &65.6±1.8 &64.6±0.5 &52.7±0.5& 49.7±0.3  \\
    &GraphLoG & ICML'21 &73.5±0.7&58.6±0.4 &62.9±0.4 & 56.9±0.4 & 49.3±0.3& 58.8±0.5 & 54.8±0.5 & 42.6±0.3 & 66.8±1.7 & 69.0±1.3  \\
    &GraphMAE & KDD'22 & 74.7±0.1 & 53.3±0.1 & 66.8±0.3 & 52.8±0.8 & 43.3±0.9  & 50.9±1.4 & 51.2±0.8 & 40.9±0.3 & 64.4±2.7 & 65.9±3.8\\
    &DSLA & NeurIPS'22 &69.3±1.0&57.8±0.5&63.3±0.3& 57.9±0.7 & 50.4±0.7 & 60.9±0.6 & 53.6±1.7 & 43.3±0.9 & 68.6±1.2 & 70.8±2.0  \\
    \midrule
    % ----- Multi Modal -----
    \multirow{8}{*}{\rotatebox{90}{\textbf{Multi-Modal}}}
    &Roberta-102M & - & 74.7±1.9 & 64.2±0.8 & 59.8±0.7 & 69.0±2.6 & 71.4±14.5 & 76.7±13.2& 65.1±19.2 & 63.7±24.6 & 67.5±5.2 & 69.9±4.9  \\
    &GPT2-87M & - & 71.0±3.4 & 61.5±1.1 & 60.6±0.3 & 74.0±8.5 & 65.4±12.9 & 81.8±25.5& 73.1±20.8 & 54.1±12.9 & 83.2±21.5 & 86.1±19.8  \\
    &MolT5&EMNLP'22 &69.9±0.8&64.7±0.9 & 55.1±0.9& 65.1±0.5 & 76.7±2.1 & 65.3±1.7 & 55.9±1.1 & 49.2±1.0 & 70.3±0.8 & 73.1±1.0 \\
    &UniMol & ICLR'23 & 76.8±0.4 & 64.6±0.2 & 65.4±0.1 & 55.8±2.8 & 50.1±5.2& 59.9±6.6 & 49.9±5.6 & 43.6±1.1 & 65.4±4.9 & 65.8±1.2  \\
    &CLOOME & NC'23 &66.7±1.8&54.2±0.9 &61.7±0.4& 64.3±0.4 & 65.2±1.5 & 75.0±2.1& 56.9±0.8 & 44.2±0.8 & 70.7±0.4 & 73.6±0.8  \\
    &InfoCORE (GE) & ICLR'24 & 79.3±0.9 & 65.3±0.2 & 60.2±0.2 & 69.9±1.2 & 79.9±3.6 & 80.3±0.9& 51.6±1.8 & 51.3±2.1 & 78.6±0.3 & 77.8±1.9  \\
    &InfoCORE (CP) & ICLR'24& 73.8±2.0&62.4±0.4 &61.1±0.2&71.0±0.6 & 74.5±4.9& 84.4±1.0 & 53.5±0.7 & 53.6±2.1 & 80.8±1.5 & 79.4±3.4  \\
    &InfoAlign & ICLR'25 & \underline{81.3±0.6} & \underline{66.4±1.1} & \underline{70.0±0.1} & \underline{49.4±0.2} & \underline{39.7±0.4} & \underline{48.4±0.6} & \underline{39.2±0.3} & \underline{40.5±0.6} & 66.7±1.7 & \underline{62.0±1.5}  \\
    
    &\textbf{CHMR} & \textbf{Ours} & {\textbf{84.7±0.2}} & {\textbf{69.3±0.3}} & {\textbf{71.4±0.2}} & {\textbf{40.9±0.3}} & {\textbf{33.7±0.4}} &  {\textbf{39.8±0.3}}&{\textbf{35.2±0.2}}& {\textbf{34.9±0.5}} & {\textbf{53.1±1.3}} &  {\textbf{48.5±0.9}}   \\
    \bottomrule
  \end{tabular}
    \captionof{table}{The performance comparison on four datasets is reported as mean±standard deviation. The \textbf{best} and \underline{second-best} average scores are highlighted.}
  \label{tab1}%
\end{table*}

\section{Experiments and Analysis}\label{sec:V}

To systematically evaluate the effectiveness and generalization ability of the proposed method, we conducted experiments addressing the following key research questions:

\textbf{RQ1.} Does our method outperform baselines on property prediction?

\textbf{RQ2.} Are its key modules effective?

\textbf{RQ3.} Does hyperparameter tuning confirm effectiveness?

\textbf{RQ4.} Are alignment and hierarchy modeling effective?

\textbf{RQ5.} How does it perform in real drug prediction?

\subsection{Experimental Setup}

\subsubsection{Dataset}
We pretrain our model on molecular and cellular datasets from multiple sources, constructed from DrugBank~\cite{wishart2018drugbank}, Cell Painting images~\cite{bray2016cell, chandrasekaran2023jump}, the JUMP-CP multi-omics platform~\cite{chandrasekaran2023jump}, and L1000 gene expression profiles~\cite{subramanian2017next}. It contains 129,592 molecules with complete structural modalities and highly incomplete external biological modalities, where some modalities are missing for over 90\% of molecules.

We evaluate our framework on nine benchmark datasets: 
ChEMBL~\cite{gaulton2012chembl}, ToxCast~\cite{richard2016toxcast}, Broad~\cite{moshkov2023predicting}, BACE, BBBP, ClinTox, SIDER, HIV~\cite{hu2020open}, and Biogen~\cite{fang2023prospective}, covering 728 prediction tasks across classification and regression settings. \textit{Details of datasets are provided in Appendix~D.1.}

\subsubsection{Baselines}

To validate the effectiveness of our method, we compare it with over 20 baseline models:
Fingerprint Models: Molecular fingerprint models \cite{rogers2010extended}, and AttentiveFP \cite{xiong2019pushing}.
Single-Modality Models: AttrMask/ContextPred/EdgePred \cite{hu2020strategies}, PretrainGNN \cite{hu2020strategies}, GraphCL \cite{you2020graph}, GROVER \cite{rong2020self}, JOAO \cite{you2021graph}, MGSSL \cite{zhang2021motif}, GraphLoG \cite{xu2021self}, GraphMAE \cite{hou2022graphmae}, DSLA \cite{kim2022graph}, MolCLR \cite{wang2022molecular}.
Molecular Multimodal Models: GEM \cite{fang2022geometry}, GraphMVP \cite{liu2022pre}, UniMol \cite{zhou2023unimol}, MolT5 \cite{edwards2022translation}, RoBERTa-102M/GPT2-87M \cite{datamol2024}, MOLEBLEND \cite{yu2024multimodal}, MOL-Mamba \cite{hu2025mol}.
Biological Multimodal Models: CLOOME \cite{sanchez2023cloome}, InfoCORE \cite{wang2023removing}, Atomas \cite{zhang2025atomas}, InfoAlign \cite{liu2025learning}. \textit{Details of baselines are provided in Appendix~D.2.}

\subsubsection{Implementation Details}

In our experiments, we carefully tune the hyperparameters for both pretraining and downstream evaluation to ensure stable convergence across diverse molecular property prediction tasks. We follow the training split strategy from \cite{liu2025learning}, using a 0.6:0.25:0.15 ratio for training, validation, and testing sets on the ChEMBL, Broad, and Biogen datasets. For the remaining molecular datasets, we adopt the scaffold-based splitting strategy, following the default 0.8:0.1:0.1 ratio used in OGB \cite{hu2020open}. For classification tasks, we evaluate performance using AUC, while for regression tasks, we use MAE as the evaluation metric. All experiments are conducted using different random seeds (ranging from 0 to 4) and run five times, reporting the mean and standard deviation of the results. \textit{The implementation details can be found in Appendix D.3.}

\subsection{Performance Evaluation}
\subsubsection{Molecular Property Prediction Comparison Results (\textbf{RQ1})}
To validate the effectiveness of our method, we conduct comparative experiments on multiple molecular property and function prediction tasks. Table \ref{tab1} presents the performance of our method on four representative drug molecule prediction datasets (ChEMBL, ToxCast, Broad, Biogen), where the first three are multi-task classification tasks used for drug activity prediction, biological response, or toxicity prediction, and the last one is a multi-task regression task for in vivo absorption, distribution, metabolism, and excretion (ADME) properties. The comparison methods are divided into two categories: single-modality and multimodal methods. The experimental results show that our method achieves the best performance on all metrics. For instance, on Biogen, the average MAE is reduced by 17.2\% compared to the second-best method, InfoAlign. Additionally, for classification tasks, we achieve an average improvement of approximately 2.0–4.4\%. Notably, these gains are observed under biological multimodal settings where missing modalities are common, demonstrating that our framework remains robust and effective in handling incomplete data.

To further assess performance, we report the proportion of tasks exceeding high AUC thresholds (80\%, 85\%, 90\%), showing that CHMR significantly outperforms all baselines across these confidence levels. In addition, we extend the evaluation to a broader set of benchmark datasets, including multi-label and toxicity prediction tasks. CHMR again achieves the best average performance (82.2\% AUC), surpassing InfoAlign (79.1\%) and MOL-Mamba (80.8\%), with 2.0–3.0\% gains on multi-label datasets. \textit{Detailed results can be found in Appendix~D.4.}

\begin{table}[htbp]\small
\centering
\renewcommand{\arraystretch}{0.5}
\renewcommand{\tabcolsep}{0.3mm}
\begin{tabular}{cc|cccc|c}
\toprule
\multicolumn{2}{c|}{\textbf{Model Variant}} & \textbf{ChEMBL $\uparrow$} & \textbf{ToxCast $\uparrow$} & \textbf{Broad $\uparrow$} & \textbf{Biogen $\downarrow$} & \textbf{$\Delta$} \\
\midrule
\multicolumn{7}{c}{\textbf{Ours}} \\

\textbf{a. Full Model} &  & \textbf{84.7$\pm$0.2} & \textbf{69.3$\pm$0.3} & \textbf{71.4$\pm$0.2} & \textbf{40.9$\pm$0.3} & -- \\
\midrule
\multicolumn{7}{c}{\textbf{Modality Augmentation}} \\
b. Zero          &  & 81.6$\pm$0.4 & 66.4$\pm$0.4 & 68.7$\pm$0.3 & 44.8$\pm$0.4 & $-5.3$ \\
c. Random        &  & 81.9$\pm$0.5 & 66.9$\pm$0.5 & 69.0$\pm$0.3 & 44.1$\pm$0.5 & $-4.5$ \\

d. Neighbor      &  & 83.1$\pm$0.3 & 67.2$\pm$0.3 & 70.0$\pm$0.2 & 42.8$\pm$0.3 & $-2.9$ \\
\midrule
\multicolumn{7}{c}{\textbf{Semantic Consistency Alignment}} \\
e. w/o SCA                  &  & 82.4$\pm$0.3 & 66.7$\pm$0.3 & 69.5$\pm$0.3 & 43.1$\pm$0.4 & $-3.6$ \\

f. w/o DA &  & 83.5$\pm$0.3 & 68.2$\pm$0.3 & 70.5$\pm$0.2 & 41.9$\pm$0.3 & $-1.7$ \\
g. w/o IA         &  & 82.8$\pm$0.4 & 67.0$\pm$0.4 & 70.1$\pm$0.3 & 42.6$\pm$0.4 & $-2.9$ \\
\midrule

\multicolumn{7}{c}{\textbf{Tree-VQ}} \\
h. w/o Tree-VQ                     &  & 82.3$\pm$0.3 & 66.9$\pm$0.3 & 69.0$\pm$0.3 & 43.4$\pm$0.4 & $-3.9$ \\

i. Flat VQ                   &  & 83.2$\pm$0.3 & 67.9$\pm$0.3 & 70.4$\pm$0.2 & 42.0$\pm$0.3 & $-2.0$ \\
\midrule
\multicolumn{7}{c}{\textbf{Context-Propagation Reconstruction}} \\
j. w/o CPR                    &  & 82.6$\pm$0.3 & 66.8$\pm$0.3 & 69.4$\pm$0.3 & 43.0$\pm$0.4 & $-3.5$ \\

k. w/o Walk      &  & 83.3$\pm$0.3 & 68.0$\pm$0.3 & 70.3$\pm$0.2 & 42.1$\pm$0.3 & $-2.0$ \\
\midrule
\multicolumn{7}{c}{\textbf{Multi-Modal Synergy Verification}} \\
l. Mol-Only                  &  & 81.5$\pm$0.4 & 66.8$\pm$0.4 & 68.7$\pm$0.4 & 44.3$\pm$0.4 & $-4.9$ \\
m. Mol+Gene                  &  & 82.7$\pm$0.3 & 67.5$\pm$0.3 & 70.0$\pm$0.3 & 43.6$\pm$0.5 & $-3.4$ \\

n. Mol+Cell                  &  & 82.6$\pm$0.3 & 67.1$\pm$0.3 & 69.8$\pm$0.3 & 43.1$\pm$0.4 & $-3.3$ \\
o. Mol+Express                  &  & 82.4$\pm$0.3 & 66.9$\pm$0.3 & 69.7$\pm$0.3 & 43.3$\pm$0.4 & $-3.6$ \\
\midrule
\multicolumn{7}{c}{\textbf{SOTA Baseline}} \\

p. InfoAlign                 &  & 81.3$\pm$0.6 & 66.4$\pm$1.1 & 70.0$\pm$0.1 & 49.4$\pm$0.2 & $-7.7$ \\
\bottomrule
\end{tabular}
\caption{Ablation study of our framework. The $\Delta$ reports the average relative change (\%) vs. Full Model.}
\label{tab:ablation}
\end{table}

\subsubsection{Effectiveness of Key Modules (\textbf{RQ2})}

To rigorously assess the contribution of each component in our framework, we conduct ablation studies (Table \ref{tab:ablation}). For MA, naive zero imputation (variant b) and random imputation (c) significantly reduce performance by $5.3\%$ and $4.5\%$, respectively, while neighborhood-based imputation (d) results in a smaller drop by $2.9\%$, indicating that leveraging neighborhood information through graph propagation yields more semantically coherent reconstructions than simple imputation strategies. Removing SCA (e) leads to a notable performance decrease by $3.6\%$, highlighting the importance of aligning augmented and original modality distributions. Ablating either distribution-level alignment (f) or instance-level alignment (g) alone also harms performance by $1.7\%$ and $2.9\%$, respectively, demonstrating their complementary roles in mitigating inconsistencies during imputation. Similarly, excluding Tree-VQ (h) or replacing Tree-VQ with flat vector quantization (i) leads to performance degradation by $3.9\%$ and $2.0\%$, respectively, indicating that hierarchical quantization better captures modality semantics and cross-modal relationships. For CPR, excluding the reconstruction loss (j) or disabling random walks in the biological graph (k) results in a performance decrease by $3.5\%$ and $2.0\%$, respectively, confirming the benefit of contextual supervision in regularizing missing modalities. Finally, testing multi-modal synergy, using molecular features alone (l) leads to substantial performance drops by $4.9\%$, while combining modalities such as Mol+Gene (m), Mol+Cell (n), or Mol+Gene Expression (o) partially recovers performance, validating the benefit of integrating complementary biological information. Overall, our full model consistently outperforms all ablations and achieves superior results compared to the SOTA baseline, InfoAlign (p).

\begin{figure}[t]
    \centering
    \subfigure[$\lambda_1$ and $\lambda_2$ on ChEMBL]{
        \includegraphics[width=0.22\textwidth]{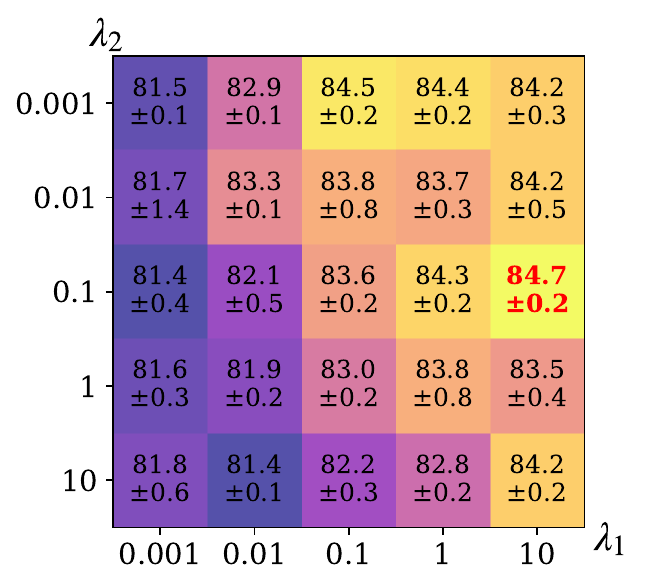}
    }
    \subfigure[$\lambda_1$ and $\lambda_2$ on Biogen]{
        \includegraphics[width=0.22\textwidth]{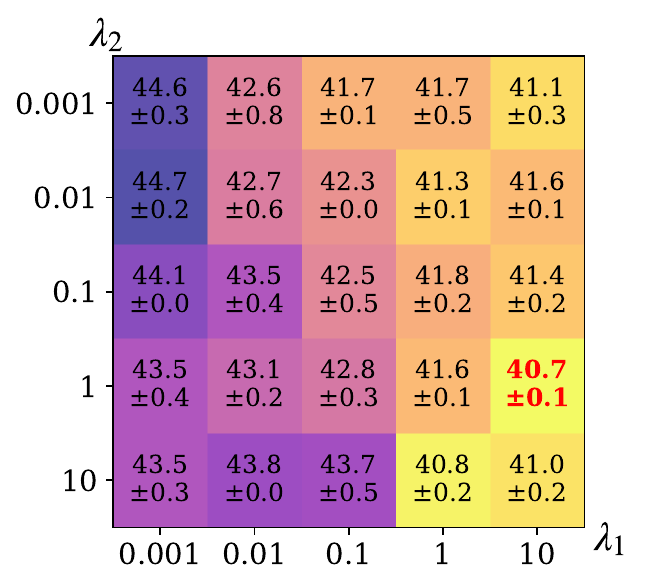}
    }
    \subfigure[Hyperparameter $\eta$]{
        \includegraphics[width=0.22\textwidth]{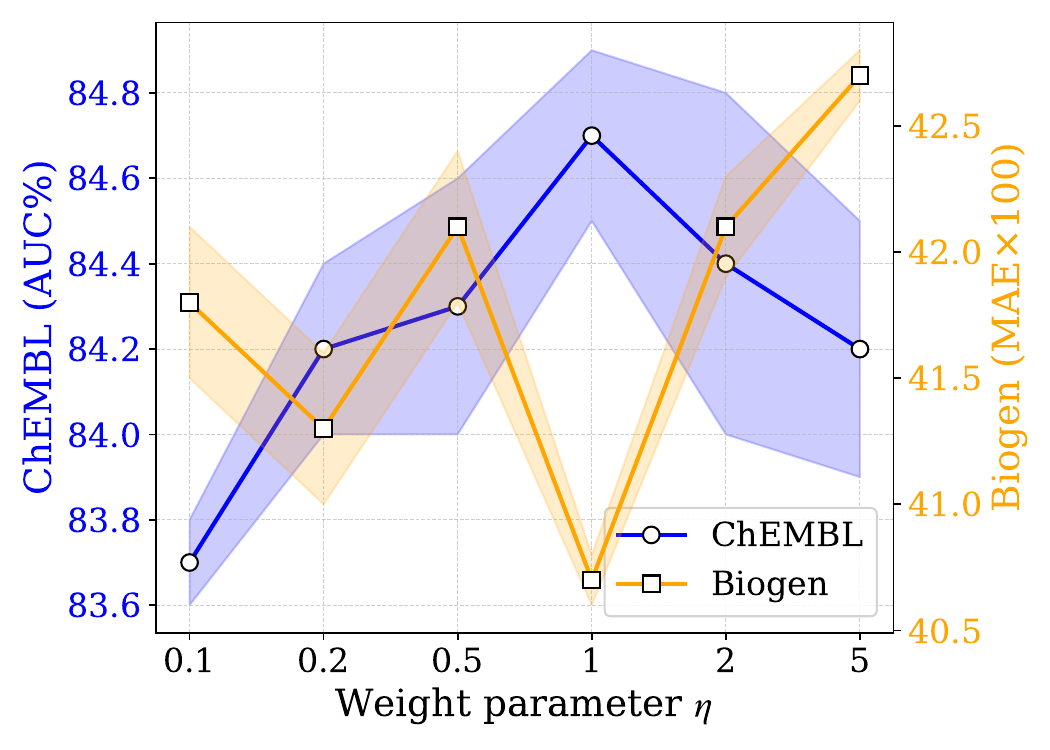}
    }
    \subfigure[Hyperparameter $h$]{
        \includegraphics[width=0.22\textwidth]{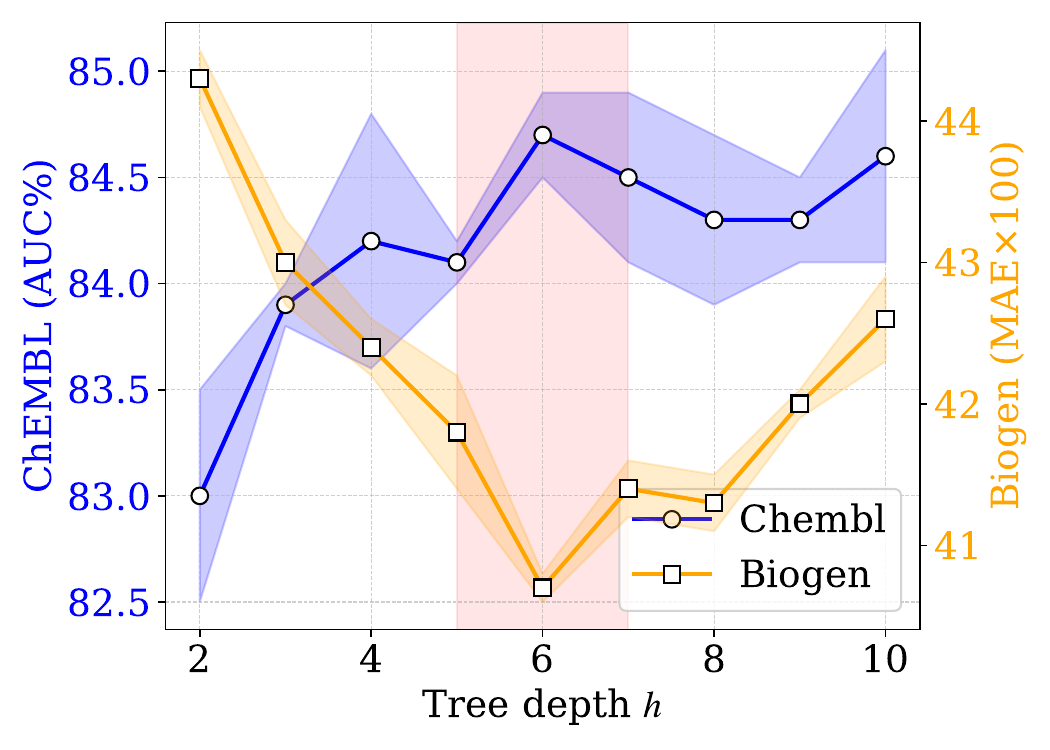}
    }
    \caption{Sensitivity analysis of hyperparameters on ChEMBL (AUC\% $\uparrow$) and Biogen (MAE $\times 100 \downarrow$).}
    \label{fig:hyperparams}
\end{figure}

\begin{figure*}[htpb]
    \centering
    \includegraphics[width=1\textwidth]{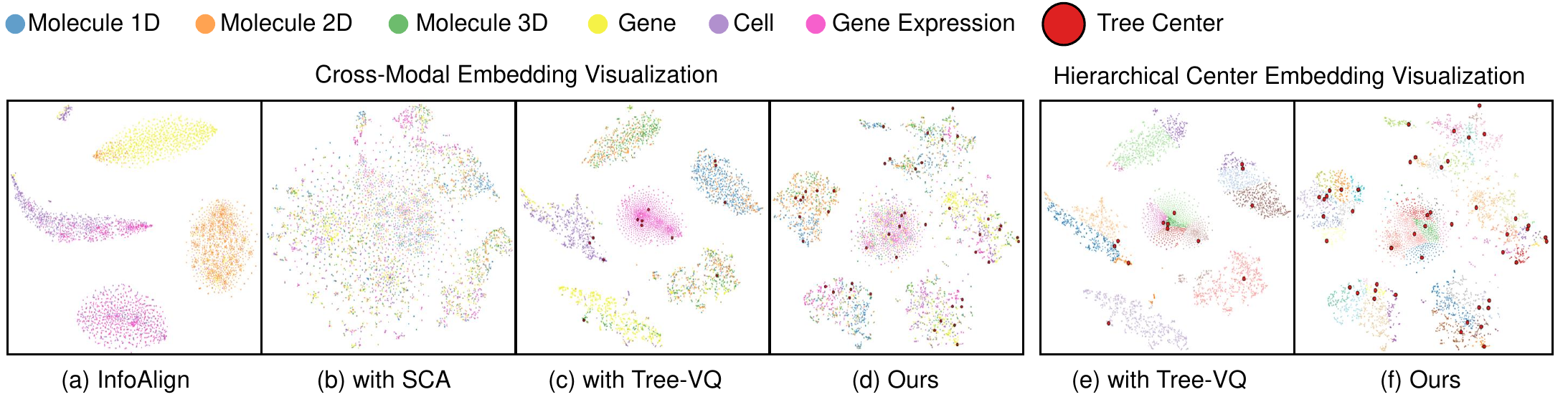}
    \caption{Visualization of cross-modal alignment and hierarchy. (a–d) show embeddings colored by modality; (e–f) display Tree codes with color-coded assignments and red dots indicating active hierarchical centers.}
    \label{fig4}
\end{figure*}

\subsubsection{Robustness across Hyperparameters (\textbf{RQ3})}

We evaluate the impact of key hyperparameters on model performance, including the loss balancing coefficients $\lambda_1$, $\lambda_2$, the weight parameter $\eta$, and the tree depth $h$. Figures \ref{fig:hyperparams}(a)–(b) show grid search results over $\lambda_1, \lambda_2 \in \{0.001, 0.01, 0.1, 1, 10\}$. $\lambda_1$, which controls the weight of the SCA module, shows that small values (\emph{e.g.}, $0.001$ or $0.01$) fail to enforce sufficient semantic consistency across modalities, leading to performance degradation. In contrast, larger values (\emph{e.g.}, $\lambda_1{=}10$) yield the best results on both ChEMBL and Biogen datasets. $\lambda_2$ governs the strength of the Tree-VQ module. Moderate values (\emph{e.g.}, $0.1$ or $1$) achieve optimal performance, indicating that appropriate hierarchical structural constraints help capture the latent dependencies among molecules, cells, and genes. However, overly large $\lambda_2$ values may impose excessive constraints and limit flexibility. Figure \ref{fig:hyperparams}(c) investigates $\eta \in \{0.1, 0.2, 0.5, 1, 2, 5\}$, which controls the commitment weight between the encoder and tree nodes in Tree-VQ. The model achieves peak performance at $\eta{=}1$, suggesting that balanced bidirectional commitment effectively aligns the projections with the discrete semantic paths. Finally, Figure \ref{fig:hyperparams}(d) examines the effect of tree depth $h \in \{2,3,\dots,10\}$. Shallow trees ($h{\leq}4$) exhibit limited capacity and underperform, while very deep trees ($h{\geq}8$) may cause overfitting and semantic fragmentation. A moderate depth ($h{=}6$) strikes a good trade-off between expressiveness and generalization.

\subsubsection{Cross-Modal Alignment and Hierarchical Representation Analysis (\textbf{RQ4})}
Figure \ref{fig4} presents two sets of t-SNE visualizations to evaluate the effectiveness of CHMR in aligning multi-modal features and capturing hierarchical dependencies. In the Cross-Modal Embedding Visualization (Figure \ref{fig4}(a–d)), we compare four methods in terms of modality alignment. Figure \ref{fig4}(a) (InfoAlign) shows that the four modalities remain largely separated, indicating poor alignment. Figure \ref{fig4}(b) (SCA only) achieves better alignment across modalities, but the resulting clusters are flat and lack hierarchical organization. Figure \ref{fig4}(c) (Tree-VQ only) captures hierarchical clusters but fails to align modalities, with each modality still forming disjoint clusters. In contrast, Figure \ref{fig4}(d) (CHMR) demonstrates both strong cross-modal alignment and clear hierarchical structuring, where modalities of the same sample overlap in the latent space and clusters are organized in a multi-level hierarchy. In the Hierarchical Center Embedding Visualization (Figure \ref{fig4}(e–f)), we further examine the Tree codes learned by different methods. Figure \ref{fig4}(e) shows that Tree-VQ alone utilizes only a limited number of Tree codes, resulting in poorly differentiated clusters. Figure \ref{fig4}(f) (CHMR) achieves full utilization of Tree codes and generates compact clusters with clear hierarchical branching, demonstrating its ability to capture fine-grained semantic dependencies across modalities. \textit{In addition, we provide full-level visualizations of the hierarchical centers learned by Tree-VQ and CHMR in Appendix D.5.}

\subsubsection{Real-World Drug Property Prediction (\textbf{RQ5})}
\begin{figure}[t]
    \centering
    \includegraphics[width=0.49\textwidth]{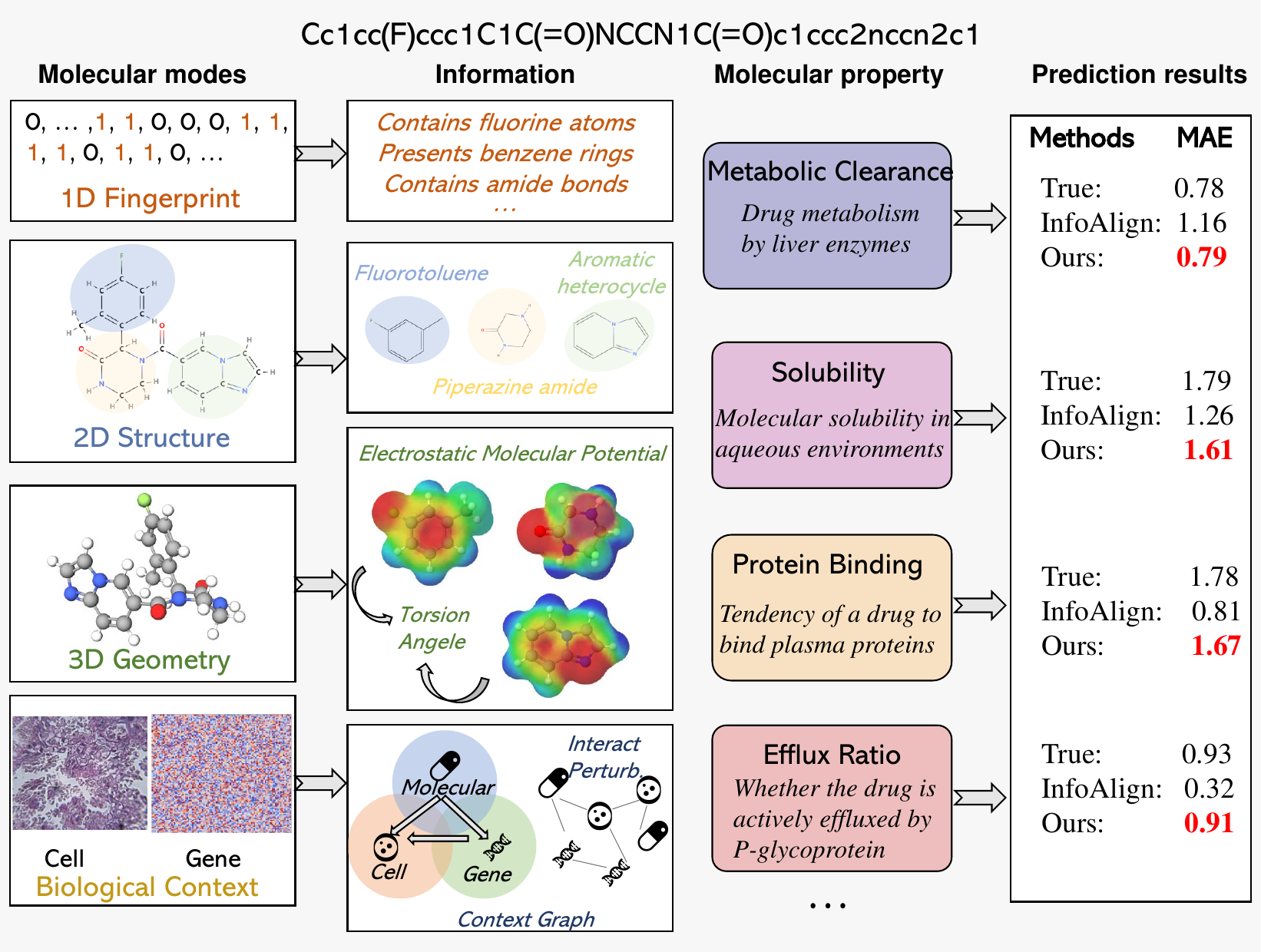}
    \caption{A case study illustrating CHMR's multi-level representation learning. Molecular 1D/2D/3D structures and biological modalities jointly provide complementary cues for pharmacological prediction. CHMR outperforms InfoAlign in prediction accuracy on the majority of tasks.}
    \label{fig5}
\end{figure}

To further demonstrate the interpretability and effectiveness of our model, we present a case study (Figure~\ref{fig5}) based on real-world drug property prediction. Given a compound, we decompose it into four hierarchical modalities: 1D fingerprints, representing substructure patterns; 2D topology, highlighting key molecular motifs; 3D conformation, encoding geometric and electrostatic features; and external biological context, including cell morphology and gene expression responses. Each modality provides complementary insights at distinct biological levels. For instance, 1D fingerprints capture substructures such as fluorotoluene, which are particularly relevant to metabolic clearance. Meanwhile, 3D spatial conformations directly affect protein binding affinity. Furthermore, the biological context dimension, which simulates cellular responses and multi-omics effects, plays a critical role in improving the prediction of P-glycoprotein (P-gp)-mediated drug efflux. The predicted pharmacological properties, shown on the right, demonstrate that our approach outperforms InfoAlign in predicting four key pharmacological endpoints, achieving lower prediction errors. This highlights the advantages of multi-level semantic alignment in improving the accuracy and robustness of drug property prediction, especially when integrating molecular structure with cellular and genetic data.

\section{Conclusion}\label{sec:VI}
We propose CHMR, a unified framework for cell-aware hierarchical multi-modal representation learning that integrates molecular structures with cellular and genomic responses. To address modality incompleteness and hierarchical dependencies, CHMR incorporates structure-aware propagation, semantic consistency alignment, context-guided reconstruction, and tree-structured vector quantization for robust cross-modal representation learning. Extensive experiments on nine benchmark datasets covering 728 molecular property prediction tasks demonstrate consistent and significant improvements over state-of-the-art baselines in both classification and regression. These results underscore CHMR’s robustness to missing modalities and its ability to capture cross-scale biological mechanisms, offering a promising direction for biologically grounded molecular modeling and drug discovery.

\section*{Acknowledgments}
This work was supported by the Scientific Research Innovation Capability Support Project for Young Faculty (No. ZYGXQNJSKYCXNLZCXM-I28), the Tongchuang Intelligent Medical Inter-disciplinary Talent Training Fund of Sun Yat-sen University (No. 76160-54990001), the National Natural Science Foundation of China (No. U21A20427), the National Key R\&D Program of China (No. 2022ZD0115100), and the Center of Synthetic Biology and Integrated Bioengineering of Westlake University (No. WU2022A009), the Zhejiang Province Selected Funding for Postdoctoral Research Projects (No. ZJ2025113).

We thank the Westlake University HPC Center for providing computational resources. This work was supported by the InnoHK program. We thank Prof. Zhen Lei from the Center for Artificial Intelligence and Robotics (CAIR), Hong Kong Institute of Science and Innovation (HKISI) for his valuable suggestions and comments. We additionally thank Bo Li, a Ph.D. candidate at the University of Macau, for his helpful assistance during this work.
\bibliography{aaai2026}

\newpage
\appendix

\section{Related Work}\label{sec:related_work}

\subsection{Molecular Representation Learning}
In recent years, molecular representation learning has achieved significant progress in drug discovery and chemical property modeling. Early approaches mainly relied on fixed molecular fingerprints (\emph{e.g.}, ECFP)~\cite{rogers2010extended} or handcrafted descriptors~\cite{krenn2020self}. With the development of graph neural networks (GNNs)~\cite{kipf2016semi, kearnes2016molecular, coley2017convolutional}, many studies have focused on learning data-driven representations over molecular graphs. D-MPNN~\cite{yang2019analyzing} introduced “directed bonds” as the center of message passing to avoid redundant propagation and loop noise in atom-centric methods, while incorporating RDKit descriptors to enhance global structural awareness. Hu et al.~\cite{hu2020strategies} proposed a hybrid pre-training strategy combining node-level and graph-level objectives, including context prediction, attribute masking, and multi-task supervision, to improve transferability on low-resource tasks. For structural semantics, GROVER~\cite{rong2020self} combined GNNs with Transformers to capture long-range dependencies, using motif and property prediction to enhance structural granularity. MolCLR~\cite{wang2022molecular} adopted contrastive learning with chemical semantic augmentations such as atom masking, bond deletion, and subgraph perturbation, enabling structure-sensitive representations from large-scale unlabeled data.

To address structural heterogeneity and distribution shifts, TopExpert~\cite{kim2023learning} proposed a topology-specific expert framework, dynamically routing molecules with diverse topologies to specialized sub-models for better generalization on scaffold shift and out-of-distribution (OOD) scenarios. ASE-Mol~\cite{jiang2025adaptive} introduced BRICS decomposition and an expert architecture to improve recognition of positive and negative substructures. APMP~\cite{jia2025association} focused on extracting high-confidence atomic path patterns through random walks and confidence filtering, coupled with Transformer aggregation to build task-specific pattern libraries. Wan et al.~\cite{wan2025multi} presented a prompt-guided pre-training and fine-tuning framework that integrates multi-channel structural learning and prompt tokens, improving combinatorial reasoning and task adaptability.

However, these methods focus mainly on single-structure modalities and lack the ability to model multi-modal signals, making it difficult to extend to molecule–cell level information in complex biological systems.

\subsection{Multi-Modal Molecular Representation}
To better capture molecular properties, recent studies have explored integrating diverse modalities such as SMILES, 2D graphs, 3D conformations, biological responses, and molecular language into unified frameworks. In 3D geometry, GraphMVP~\cite{liu2022pre} enabled 2D-to-3D semantic transfer via contrastive learning and generative reconstruction, without requiring 3D inputs during inference. GEM~\cite{fang2022geometry} and Uni-Mol~\cite{zhou2023unimol} introduced geometry-level networks and SE(3)-equivariant Transformers to improve the modeling of 3D conformations.

Holi-Mol~\cite{kim2024holistic} proposed a 2D–3D fusion framework with cross-view contrastive learning, but its modality scheduling and task-awareness remain static. PremuNet~\cite{zhang2024pre} combined SMILES, 2D graphs, and 3D conformations with dual-branch architectures to extract complementary features. MOLEBLEND~\cite{yu2024multimodal} introduced an atom-relation level fusion mechanism for fine-grained structural alignment. MOL-Mamba~\cite{hu2025mol} incorporated electronic semantics at the atomic level, jointly learning with structural information in a graph state space model. MolGT~\cite{chen2025pretraining} unified 2D and 3D encoding with a graph Transformer and multi-task objectives for fine-grained alignment. Despite these advances, most approaches assume complete multimodal data and are vulnerable to strong modality dominance when weak modalities are noisy or missing.

For molecular language models, MolT5~\cite{edwards2022translation} and ChemGPT~\cite{frey2023neural} established general-purpose chemical language models, enabling cross-modal translation between molecules and text. Mol-CA~\cite{liu2023molca} projected structural graphs into large language model (LLM) spaces with a Q-Former, supporting cross-modal generation and retrieval. Atomas~\cite{zhang2025atomas} introduced hierarchical alignment from atoms, fragments, and molecules to text for fine-grained structure–text mapping.

While these methods improve multi-modal fusion, they do not explicitly model the hierarchical dependencies between molecules, cells, and genes, making it hard to capture cross-scale biological mechanisms.

\subsection{Cell-Responsive Multi-Modal Molecular Representation}
With the rise of large-scale cell imaging and multi-omics data, recent studies have explored incorporating cellular-level phenotypic responses into molecular modeling to address the semantic gap in functional biology. MoCoP~\cite{nguyen2023molecule} first used cell phenotype images as functional supervision in molecular pre-training, aligning structure and phenotype through contrastive learning and significantly enhancing biological semantic awareness. CLOOME~\cite{sanchez2023cloome} extended this to bidirectional retrieval tasks with structure–phenotype paired contrastive loss (InfoLOOB) and a shared embedding space. MolPhenix~\cite{fradkin2024molecules} proposed a novel phenotype-to-structure paradigm for function-driven reverse inference.

Further expanding modality dimensions, InfoCORE~\cite{wang2023removing} focused on removing non-biological batch effects by maximizing conditional mutual information for bias-robust structure–phenotype modeling. InfoAlign~\cite{liu2025learning} and Masood et al.~\cite{masood2025multi} integrated molecular structures, cellular morphology (Cell Painting), and transcriptomics (L1000) into a shared representation space, aligning molecules with their biological responses to enrich both chemical and functional semantics and improve generalization on pharmacological and toxicity prediction tasks.

However, existing approaches still face key limitations. External biological modalities (\emph{e.g.}, gene expression or cell morphology) are often incomplete or missing, leading to modality imbalance and degraded generalization. Most methods assume complete multimodal input and perform static fusion or flattened alignment, making them vulnerable to modality missingness and distribution shifts.

\begin{table*}[htpb]\small
  \centering
  \begin{tabular}{p{3cm} p{10cm}}
    \toprule
    \textbf{Symbol} & \textbf{Description} \\
    \midrule
    $\mathcal{V}=\{v_i\}_{i=1}^N$ & Molecular sample set, with $N$ representing the total number of molecules. \\
    $v_i$ & The $i$-th molecule sample. \\
    $\xi\in\mathcal{M}\cup\mathcal{C}$ & Generalized modality index. \\
    $m \in \mathcal{M}$ & Molecular modality index. \\
    $c \in \mathcal{C}$ & External biological modality index. \\
    $\mathbf{X}_i=\{\mathbf{x}_i^{\xi}\}_{\xi\in\mathcal{M}\cup\mathcal{C}}$ & Multi-modal feature set of molecule $v_i$. \\
    $\mathbf{x}_i^{m}\in\mathbb{R}^{d_m}$ & Feature vector for molecular modality $m$, with $d_m$ as the feature dimension. \\
    $\mathbf{x}_i^{c}\in\mathbb{R}^{d_c}$ & Feature vector for biological modality $c$, with $d_c$ as the feature dimension. \\
    $\mathcal{C}_i^{\mathrm{obs}}\subseteq\mathcal{C}$ & Observed external biological modalities for molecule $v_i$. \\
    $\mathcal{C}_i^{\mathrm{miss}}=\mathcal{C}\setminus\mathcal{C}_i^{\mathrm{obs}}$ & Missing external biological modalities for molecule $v_i$. \\
    \midrule
    \multicolumn{2}{c}{\textbf{Modality Augmentation}} \\
    \midrule
    $\mathbf W\in\mathbb{R}^{N\times N}$ & Molecular sample similarity weight matrix. \\
    $\mathbf W_{ij}$ & Similarity weight between molecules $v_i$ and $v_j$ in the structural space. \\
    $\mathcal{N}_K(v_i)$ & Top $K$ nearest neighbors for molecule $v_i$. \\
    $T$ & Number of iterations for feature propagation. \\
    $\mathbf{x}_i^{c,\text{aug}}$ & Augmented external biological modality feature after feature propagation. \\
    \midrule
    \multicolumn{2}{c}{\textbf{Semantic Consistency Alignment}} \\
    \midrule
    $f^{m}(\cdot)$ & Projection function for molecular modality $m$. \\
    $f^{c}(\cdot)$ & Projection function for external biological modality $c$. \\
    $\mathbf{p}_i^{m}$ & Projection vector for molecular modality of molecule $v_i$. \\
    $\mathbf{p}_i^{c}$ & Projection vector for external biological modality of molecule $v_i$. \\
    $\mathbf{p}_i^{c,\text{aug}}$ & Projection vector for augmented external biological modality of molecule $v_i$. \\
    $\mathbf{p}_i^{\text{anch}}$ & Aggregated projection vector for all molecular modalities of molecule $v_i$. \\
    \midrule
    \multicolumn{2}{c}{\textbf{Tree-Structured Vector Quantization}} \\
    \midrule
    $\mathcal{T}=\bigcup_{h=1}^H\mathcal{E}^{h}$ & Binary tree structure with depth $H$. \\
    $\mathcal{E}^{h}=\{\mathbf{e}_j^{h}\}_{j=1}^{2^h}$ & Set of nodes at the $h$-th level of the binary tree. \\
    $\mathbf{e}_j^{h}$ & $j$-th node embedding at the $h$-th level of the binary tree. \\
    $\delta_j^{\xi,h}$ & Cosine distance between modality projection vector $\mathbf{p}^{\xi}$ and tree node $\mathbf{e}_j^{h}$. \\
    $\tilde{\delta}_j^{\xi,h}$ & Distance after applying the tree routing mask. \\
    $j_{\mathrm{par}}^{\xi,h-1}$ & Parent node index for modality $\xi$ at the $(h-1)$-th level. \\
    $j^{*\xi,h}$ & Quantized node index for modality $\xi$ at the $h$-th level. \\
    $\mathbf{q}^{\xi,h}$ & Quantized vector for modality $\xi$ at the $h$-th level. \\
    \midrule
    \multicolumn{2}{c}{\textbf{Context-Propagation Reconstruction}} \\
    \midrule
    $\mathcal{H}=(\mathcal{U},\mathcal{R})$ & Biological context graph: node set $\mathcal{U}$ and edge set $\mathcal{R}$. \\
    $u_i$ & Node in graph $\mathcal{H}$, representing either a molecule or external biological modality. \\
    $(u_i,u_j)\in\mathcal{R}$ & An edge between nodes $u_i$ and $u_j$ in the context graph. \\
    $\beta(u_i,u_j)\in[0,1]$ & Normalized weight of edge $(u_i,u_j)$ in the context graph. \\
    $L$ & Maximum path length for context propagation random walk. \\
    $(u_i,u_{i_1},...,u_{i_L})$ & Random walk path starting from node $u_i$. \\
    $\beta_{i,l}$ & Cumulative propagation weight for the $l$-th node in the random walk path. \\
    $g^{m}(\cdot)$ & Decoder for molecular modality $m$. \\
    $g^{c}(\cdot)$ & Decoder for external biological modality $c$. \\
    $\hat{\mathbf{x}}_i^{m}$ & Reconstructed molecular modality feature. \\
    $\hat{\mathbf{x}}_i^{c}$ & Reconstructed external biological modality feature. \\
    \midrule
    \multicolumn{2}{c}{\textbf{Loss Functions and Hyperparameters}} \\
    \midrule
    $\lambda_1,\lambda_2$ & Hyperparameters controlling the weights of $\mathcal{L}_{\mathrm{CPR}}$ and $\mathcal{L}_{\mathrm{TreeVQ}}$. \\
    $\eta$ & Weight hyperparameter for reverse commitment in the $\mathcal{L}_{\mathrm{TreeVQ}}$. \\
    $H$ & Maximum depth of the binary tree. \\
    \bottomrule
  \end{tabular}
    \caption{Symbol table for CHMR framework.}
\end{table*}

\section{Theoretical Analysis} \label{sec:theoretical_analysis}

\subsection{Theoretical Interpretation of Feature Propagation}

The proposed modality augmentation strategy can be theoretically interpreted as an iterative solution to a constrained Dirichlet energy minimization problem \cite{rossi2022unreasonable} over the molecular similarity graph.

Let $\mathbf{x}^{c} = [\mathbf{x}_1^c, \dots, \mathbf{x}_N^c]^\top \in \mathbb{R}^{N \times d}$ denote the feature vector of modality $c$ across all $N$ molecules. We define the Dirichlet energy over the molecular similarity graph $\mathcal{G} = (\mathcal{V}, \mathbf{W})$ as:
\begin{equation}
\mathbb{E}(\mathbf{x}^{c}) = \frac{1}{2} \sum_{i,j} \mathbf{W}_{ij} \left\| \mathbf{x}_i^c - \mathbf{x}_j^c \right\|^2 = \mathrm{Tr}\left( \mathbf{x}^{c\top} \mathbf{L} \mathbf{x}^c \right),
\end{equation}
where $\mathbf{L} = \mathbf{D} - \mathbf{W}$ is the graph Laplacian, and $\mathbf{D}$ is the diagonal degree matrix with $\mathbf{D}_{ii} = \sum_j \mathbf{W}_{ij}$.

We aim to find a smoothed estimate $\tilde{\mathbf{x}}^{c}$ that minimizes $\mathbb{E}(\mathbf{x}^c)$ under the constraint that observed features remain fixed:
\begin{equation}
\min_{\hat{\mathbf{x}}^c} \quad \mathbb{E}(\tilde{\mathbf{x}}^c) \quad \text{subject to} \quad \tilde{\mathbf{x}}_i^c = \mathbf{x}_i^c, \quad \forall i \in \mathcal{C}_{\mathrm{obs}}.
\end{equation}

This constrained optimization can be solved iteratively using a local propagation rule:
\begin{equation}
\tilde{\mathbf{x}}_i^{c,(T)} =
\begin{cases}
\mathbf{x}_i^c, & i \in \mathcal{C}_{\mathrm{obs}}, \\[1ex]
\sum\limits_{j \in \mathcal{N}_K(v_i)} \mathbf{W}_{ij} \tilde{\mathbf{x}}_j^{c,(T-1)}, & i \in \mathcal{C}_{\mathrm{miss}},
\end{cases}
\end{equation}

This process gradually minimizes the energy $\mathbb{E}(\tilde{\mathbf{x}}^c)$ while satisfying the hard constraints on observed entries, and corresponds to solving a Laplacian system with Dirichlet boundary conditions. It is guaranteed to converge to a unique minimum under mild assumptions. Therefore, our augmentation strategy can be viewed as a principled graph-based imputation method that preserves observed data while enforcing smoothness over structurally similar molecules.

\subsection{VICReg as Feature Distribution Alignment}

Let $\mathbf{Z} = [\mathbf{z}_1^\top, \ldots, \mathbf{z}_N^\top]^\top \in \mathbb{R}^{N \times d}$, $\mathbf{Z}' = [\mathbf{z}_1'^\top, \ldots, \mathbf{z}_N'^\top]^\top \in \mathbb{R}^{N \times d}$, where $\mathbf{z}_i = \mathbf{p}_i^c$, $\mathbf{z}_i' = \mathbf{p}_i^{c,\text{aug}}$.

VICReg loss \cite{bardes2022vicreg} is defined as:
\begin{equation}
\begin{aligned}
\mathrm{VICReg}(\mathbf{Z}, \mathbf{Z}') &=  \underbrace{\frac{1}{N} \sum_{i=1}^N \| \mathbf{z}_i - \mathbf{z}_i' \|_2^2}_{\text{Invariance}} \\
&+ \underbrace{\sum_{j=1}^d \max\left(0, 1 - \mathrm{Var}(\mathbf{Z}_{:,j})\right)}_{\text{Variance}} \\
&+  \underbrace{\| \mathrm{Cov}(\mathbf{Z}) - \mathbf{I} \|_F^2}_{\text{Covariance}},
\end{aligned}
\end{equation}
where \begin{align}
\mathrm{Var}(\mathbf{Z}_{:,j}) &= \frac{1}{N} \sum_{i=1}^N \left(\mathbf{z}_{ij} - \frac{1}{N} \sum_{k=1}^N \mathbf{z}_{kj}\right)^2, \\
\mathrm{Cov}(\mathbf{Z}) &= \frac{1}{N} \mathbf{Z}^\top \mathbf{Z} - \left(\frac{1}{N} \sum_{i=1}^N \mathbf{z}_i\right)\left(\frac{1}{N} \sum_{i=1}^N \mathbf{z}_i\right)^\top.
\end{align}

\subsection{Tree-VQ Captures Hierarchical Structure}

Let $\mathcal{P}(\mathbf{p}) = \{j_1, j_2, \dots, j_H\}$ be the quantization path obtained via level-wise routing in the binary tree, where each $j_h = \arg\min_{j \in \{2j_{h-1}, 2j_{h-1}+1\}} \delta(\mathbf{p}, \mathbf{e}_j^h)$, with $\delta(\mathbf{p}, \mathbf{e}) = 1 - \cos(\mathbf{p}, \mathbf{e})$ representing the angular distance metric.

We formally define the induced partition of the latent space $\mathbb{R}^d$ as:
\begin{equation}
\begin{aligned}
\mathbb{R}^d &= \bigcup_{j_1=1}^{2} \mathcal{R}_{j_1}^1, \\
\mathcal{R}_{j_1}^1 &= \bigcup_{j_2 \in \{2j_1, 2j_1+1\}} \mathcal{R}_{j_2}^2, \\
&\vdots \\
\mathcal{R}_{j_H}^H &= \left\{ \mathbf{p} \in \mathbb{R}^d : \mathcal{P}(\mathbf{p}) = \{j_1, \dots, j_H\} \right\},
\end{aligned}
\end{equation}
where each $\mathcal{R}_{j_h}^h$ represents the region at level $h$ associated with node $j_h$.

\textbf{Theorem.} If two inputs $\mathbf{p}_1, \mathbf{p}_2 \in \mathbb{R}^d$ follow identical paths $\mathcal{P}(\mathbf{p}_1) = \mathcal{P}(\mathbf{p}_2)$ in the Tree-VQ structure, then their final quantized vectors are identical ($\mathbf{q}_H^{(1)} = \mathbf{q}_H^{(2)} = \mathbf{e}_{j_H}^H$), and their cosine similarity satisfies:
\begin{equation}
\cos(\mathbf{p}_1, \mathbf{p}_2) \geq 2\alpha^2 - 1,
\end{equation}
where $\alpha = \cos(\mathbf{p}_1, \mathbf{q}_H) = \cos(\mathbf{p}_2, \mathbf{q}_H)$ denotes the similarity between each input and their shared quantized vector.

\textit{Proof.} Since $\mathcal{P}(\mathbf{p}_1) = \mathcal{P}(\mathbf{p}_2)$, both inputs are quantized to the same terminal node at depth $H$, yielding identical quantized vectors $\mathbf{q}_H^{(1)} = \mathbf{q}_H^{(2)} = \mathbf{e}_{j_H}^H$.

Let $\theta(\mathbf{a}, \mathbf{b}) = \arccos(\cos(\mathbf{a}, \mathbf{b}))$ denote the angular distance between vectors $\mathbf{a}$ and $\mathbf{b}$. By definition, we have:
\begin{equation}
\theta(\mathbf{p}_1, \mathbf{q}_H) = \arccos(\alpha), \quad \theta(\mathbf{p}_2, \mathbf{q}_H) = \arccos(\alpha).
\end{equation}

Applying the spherical triangle inequality in the unit hypersphere:
\begin{equation}
\theta(\mathbf{p}_1, \mathbf{p}_2) \leq \theta(\mathbf{p}_1, \mathbf{q}_H) + \theta(\mathbf{q}_H, \mathbf{p}_2) = 2\arccos(\alpha).
\end{equation}

Since the cosine function is monotonically decreasing on $[0, \pi]$, we obtain:
\begin{equation}
\cos(\theta(\mathbf{p}_1, \mathbf{p}_2)) \geq \cos(2\arccos(\alpha)).
\end{equation}

Using the double-angle trigonometric identity $\cos(2x) = 2\cos^2(x) - 1$, we derive:
\begin{equation}
\cos(2\arccos(\alpha)) = 2\cos^2(\arccos(\alpha)) - 1 = 2\alpha^2 - 1.
\end{equation}

Therefore, we establish the lower bound:
\begin{equation}
\cos(\mathbf{p}_1, \mathbf{p}_2) \geq 2\alpha^2 - 1.
\end{equation}

To verify the tightness of this bound, consider the case where $\mathbf{p}_1$ and $\mathbf{p}_2$ are symmetrically positioned around $\mathbf{q}_H$ in the plane they span. Let $\mathbf{q}_H = (1, 0, 0, \ldots, 0)$, $\mathbf{p}_1 = (\alpha, \sqrt{1-\alpha^2}, 0, \ldots, 0)$, and $\mathbf{p}_2 = (\alpha, -\sqrt{1-\alpha^2}, 0, \ldots, 0)$. Then:
\begin{equation}
\cos(\mathbf{p}_1, \mathbf{p}_2) = \alpha^2 - (1-\alpha^2) = 2\alpha^2 - 1,
\end{equation}
which achieves equality in the bound, confirming its tightness.

\textbf{Corollary 1.} For any $\mathbf{p}_1, \mathbf{p}_2$ sharing the same quantization path with $\alpha \geq \frac{\sqrt{2}}{2}$ (i.e., $\theta \leq \frac{\pi}{4}$), we have:
\begin{equation}
\cos(\mathbf{p}_1, \mathbf{p}_2) \geq 0,
\end{equation}
ensuring that inputs within the same partition region maintain non-negative similarity, which is crucial for meaningful semantic grouping.

\textbf{Corollary 2.} As the tree depth $H$ increases, the maximum angular diameter of regions $\mathcal{R}_{j_H}^H$ decreases, leading to tighter bounds:
\begin{equation}
\lim_{H \to \infty} \sup_{\mathbf{p}_1, \mathbf{p}_2 \in \mathcal{R}_{j_H}^H} \theta(\mathbf{p}_1, \mathbf{p}_2) = 0,
\end{equation}
which guarantees increasingly precise semantic partitioning at deeper levels.

Therefore, Tree-VQ partitions the latent space into semantically coherent, hierarchically organized regions where the angular diameter of each partition is explicitly bounded by the quantization depth and the similarity to the quantization centroid. This hierarchical structure directly models the cross-scale biological mechanisms discussed in Section 1, capturing the causal routing and multi-hop semantics from molecular fingerprints (shallow) to cellular phenotypes (deep) that are overlooked by flattened alignment approaches.

\begin{table*}[ht]
\centering
\begin{tabular}{cccc}
\toprule
\textbf{Type} & \textbf{Modality} & \textbf{Missing Rate} & \textbf{Missing Samples} \\
\midrule
Molecule 1D     & Fingerprint~\cite{rogers2010extended}      & - & - \\
\multicolumn{4}{c}{Concatenation of ECFP (radius=2) and MACCS fingerprints.} \\
\midrule
Molecule 2D     &Graph Structure~\cite{hwang2022analysis}      & - & - \\
\multicolumn{4}{c}{Atom-level graph topological features extracted by Virtual GNN.} \\
\midrule
Molecule 3D     & Spatial Conformation ~\cite{zhou2023unimol}      & - & - \\
\multicolumn{4}{c}{Pretrained 3D features generated by UniMol.} \\
\midrule
Cell     & CP-JUMP~\cite{chandrasekaran2023jump}      & 12.54\%  & 16,247 \\
\multicolumn{4}{c}{Cell Painting: morphological profiles of compound-treated cells} \\
\midrule
Cell    & CP-Bray~\cite{bray2016cell}       & 91.37\%  & 118,407\\ 
\multicolumn{4}{c}{Cell Painting: morphological profiles of compound-treated cells }        \\
\midrule
Gene    & G-CRISPR \cite{chandrasekaran2023jump}     & 99.43\%  & 128,853 \\ 
\multicolumn{4}{c}{CRISPR knockout experiments providing gene perturbation features} \\
\midrule
Gene   & G-ORF \cite{chandrasekaran2023jump}        & 99.45\%  & 128,883 \\ 
\multicolumn{4}{c}{ORF overexpression experiments providing gene perturbation features} \\
\midrule
Gene Expression    & L1000 Profiles~\cite{subramanian2017next} & 94.25\%  & 122,135 \\ 
\multicolumn{4}{c}{Expression levels of 978 landmark genes in compound-treated cells} \\
\bottomrule
\end{tabular}
\caption{Missing rates and descriptions of external biological modalities in the pretraining dataset. The dataset contains 129,592 molecules with paired cellular and genomic responses, but the external modalities are highly incomplete and asymmetric.}
\label{tab:modality_missing}
\end{table*}

\subsection{Computational and Space Complexity Analysis} 
We define the following notation for our complexity analysis, consistent with the framework design presented in Section~3. Let $N$ denote the number of molecular samples in the dataset. There are $M_m$ molecular modalities (corresponding to $|\mathcal{M}|$) and $M_c$ cellular modalities (corresponding to $|\mathcal{C}|$). The dimensionality of molecular feature representations is $d_m$, while cellular feature representations have dimension $d_c$. The shared latent space into which features are projected has dimension $d$.  The $K$-nearest neighbor graph construction uses $K$ as the number of neighbors considered for each sample. The modality augmentation process involves $T$ iterations of propagation to enhance missing modalities. The average missing modality ratio is denoted by $m$, where $0 \leq m \leq 1$, representing the proportion of missing cellular modalities across the dataset. We utilize a binary tree of depth $H$, and the context graph contains $\mathcal{R}$ edges, with random walks performed up to length $L$.

The time complexity of the CHMR framework is determined by analyzing the computational requirements of each component. The MA module involves $K$-nearest neighbor graph construction with time complexity $\mathcal{O}(NK)$ and iterative propagation for missing modalities with time complexity $\mathcal{O}(NKTmM_c d_c)$, where $m$ is the average missing modality ratio, $M_c$ is the number of cellular modalities, and $d_c$ is the cellular feature dimension. The SCA module dominates the computational cost due to the sample-level alignment using InfoNCE loss, which requires computing pairwise similarities between all samples, resulting in $\mathcal{O}(N^2 d)$ complexity, where $d$ is the shared latent space dimension. The Tree-VQ module processes each sample through $H$ tree layers for $(M_m + M_c)$ modalities with time complexity $\mathcal{O}(N(M_m + M_c)Hd)$. Finally, the CPR module performs random walks of length $L$ on a context graph with $\mathcal{R}$ edges with time complexity $\mathcal{O}(L\mathcal{R})$, and the reconstruction process has time complexity $\mathcal{O}(N(M_m d_m + M_c d_c))$. Combining these components, the total time complexity of CHMR simplifies to $\mathcal{O}(N^2 d + NKTmM_c d_c + N(M_m + M_c)Hd + L\mathcal{R})$. When processing large-scale molecular datasets ($N$ large), the $N^2 d$ term typically dominates the computational complexity, while high missing ratios ($m$) or numerous cellular modalities ($M_c$) make the $NKTmM_c d_c$ term significant.

The space complexity of CHMR is determined by several key factors. The MA module needs $\mathcal{O}(NK)$ space for the sparse similarity matrix and $\mathcal{O}(N(d_m M_m + d_c M_c))$ for storing multimodal features. The SCA module requires storage for the $N \times N$ similarity matrix used in InfoNCE loss, contributing $\mathcal{O}(N^2)$ space complexity, which is often the dominant factor. The Tree-VQ module contributes $\mathcal{O}(2^H d)$ space for tree node embeddings and $\mathcal{O}(N(M_m + M_c)H)$ for quantization results, where $H$ is the tree depth. The CPR module requires $\mathcal{O}(N + \mathcal{R})$ space for the context graph structure and $\mathcal{O}(N(M_m d_m + M_c d_c))$ for reconstruction outputs. Aggregating these components, the total space complexity simplifies to $\mathcal{O}(N^2 + N(K + d_m M_m + d_c M_c + (M_m + M_c)H) + 2^H d + \mathcal{R})$. In most scenarios, the $N^2$ term dominates the space requirements, though the $2^H d$ term becomes significant when the tree depth $H$ is large. To manage memory usage in practical implementations, negative sampling techniques can be applied to reduce the $N^2$ impact, and the tree depth $H$ can be constrained to control the exponential growth of $2^H d$.

\begin{table}[htpb]\small
    \centering
    \renewcommand{\arraystretch}{1}
    \renewcommand{\tabcolsep}{0.8mm}
    \begin{tabular}{cccccc}
        \toprule
        \multirow{2}{*}{\textbf{Dataset}} & \multirow{2}{*}{\textbf{Type}} & \multirow{2}{*}{\textbf{Molecules}} & \multicolumn{2}{c}{\textbf{Max. /Avg.}} & \multirow{2}{*}{\textbf{Tasks}} \\
        \cmidrule(lr){4-5}
        & & & \textbf{Atoms} & \textbf{Edges} & \\
        \midrule
        ChEMBL  & Classif. & 2355  & 61/23.7  & 68/25.6    & 41  \\
        ToxCast   & Classif. & 8576  & 124/18.7 & 268/38.5  & 617 \\
        Broad  & Classif. & 6567  & 74/34.1  & 82/36.8    & 32  \\
        BACE      & Classif. & 1513  & 97/34.1 & 202/73.7   & 1   \\
        BBBP      & Classif. & 2039  & 132/24.0  & 290/51.9    & 1   \\
        ClinTox   & Classif. & 1477  & 136/26.1 & 286/55.7   & 2   \\
        Sider     & Classif. & 1427  & 492/33.6 & 1010/70.7   & 27  \\
        HIV  & Classif. & 41127  & 222/25.5  & 502/54.9    & 1  \\
        Biogen  & Reg.     & 3521  & 78/23.2  & 84/25.3     & 6   \\
        \bottomrule
    \end{tabular}
    \caption{Summary of the downstream molecular data.}
    \label{datasets}
\end{table}

\begin{table*}[htbp]
  \centering
    \renewcommand{\arraystretch}{0.9}
    \renewcommand{\tabcolsep}{1mm}
    \begin{tabular}{c|cccccccc|cc}
    \toprule
    \multirow{2}[4]{*}{\textbf{Dataset}} & \multicolumn{8}{c|}{\textbf{Fine-Tuning}}                      & \multicolumn{2}{c}{\textbf{Pretraining}} \\
\cmidrule{2-10}          
          & Learning Rate    & Dropout & $\gamma$ & Hidden Size & Batch Size & Patience & Epochs &Norm & $\lambda_1$ & $\lambda_2$ \\
    \midrule
    ChEMBL      & 0.001 & 0.9   & None  & 1800  & 5120  & 50    & 300 &LayerNorm  & 10    & 0.1 \\
    ToxCast     & 0.01 & 0.8   & None  & 1800  & 5120  & 50    & 300  &LayerNorm & 1     & 0.1 \\
    Broad       & 0.002 & 0.8   & None  & 1800  & 5120  & 50    & 300 &LayerNorm  & 10    & 10 \\
    BBBP     & 0.001 & 0.5   & None  & 2400  & 5120  & 50    & 300  &LayerNorm & 10    & 10 \\
    BACE     & 0.0005 & 0.5   & 0.005 & 1800  & 16    & 5     & 100 &BatchNorm  & 0.01  & 0.1 \\
    ClinTox  & 0.005 & 0.9   & 0.01  & 2400  & 32    & 5     & 100   &LayerNorm& 0.1   & 10 \\
    SIDER    & 0.0005 & 0.2   & 0.1   & 1200  & 5120  & 20    & 30  &BatchNorm   & 10    & 10 \\
    HIV      & 0.001 & 0.8   & 0.001 & 1800  & 10240 & 50    & 300  &BatchNorm  & 10    & 10 \\
    Biogen      & 0.002 & 0.8   & None  & 1200  & 5120  & 50    & 300 &LayerNorm  & 10    & 1 \\
    \bottomrule
    \end{tabular}%
  \caption{Hyperparameter settings for pretraining and downstream fine-tuning on nine benchmark datasets.}
    \label{tab:hyperparam_settings}
\end{table*}

\begin{table*}[t]\small
  \centering
  \renewcommand{\arraystretch}{0.9}
  \renewcommand{\tabcolsep}{0.5mm}
  \vspace{0.5em}
  \begin{tabular}{c|cc|cccc|cccc|cccc}
    \toprule
    ~ &
    \multicolumn{2}{c|}{\textbf{Dataset}} &
    \multicolumn{4}{|c}{\textbf{ChEMBL~(AUC\% $\uparrow$)}} &
    \multicolumn{4}{|c}{\textbf{ToxCast~(AUC\% $\uparrow$)}}&
    \multicolumn{4}{|c}{\textbf{Broad~(AUC\% $\uparrow$)}}  \\
    &\multicolumn{2}{c|}{( \textbf{Molecule} /  \textbf{Task})} &
    \multicolumn{4}{|c}{(\textbf{2355 / 41})} &
    \multicolumn{4}{|c}{(\textbf{8576 / 617})}&
    \multicolumn{4}{|c}{(\textbf{6567 / 32})} \\
    \midrule
    \textbf{}&\textbf{Method} &\textbf{Venue} & \textbf{Avg.} & \textbf{>80\%} &> \textbf{85\%}  & > \textbf{90\%}  & \textbf{Avg.} & \textbf{>80\%} &> \textbf{85\%}  & > \textbf{90\%}   & \textbf{Avg.} & \textbf{>80\%} &> \textbf{85\%}  & > \textbf{90\%}    \\
    \midrule
    % ---- Single-Modal ----
    \multirow{13}{*}{\rotatebox{90}{\textbf{Single-Modal}}}
    &MLP &JCIM'10& 76.8±2.2 & 48.8±3.9& 34.6±6.3 & 21.9±5.7  & 57.6±1.0 & 1.6±0.3& 0.6±0.4 & 0.3±0.3  & 63.3±0.3 & 6.3±0.0 &  \underline{4.4±1.7} & \underline{3.1±0.0} \\
    &RF  &JCIM'10& 54.7±0.7 & 0.0±0.0 & 0.0±0.0 & 0.0±0.0 & 52.3±0.1 & 0.2±0.1 & 0.1±0.1 & 0.0±0.1  & 55.5±0.1 & 0.0±0.0 & 0.0±0.0 &0.0±0.0\\
    &GP  &JCIM'10& 51.0±0.0 & 0.0±0.0 & 0.0±0.0 & 0.0±0.0 & \multicolumn{4}{c}{Run out of Time} & 50.6±0.0 & 0.0±0.0 & 0.0±0.0 & 0.0±0.0 \\
    &AttrMask &  {ICLR'20}&  73.9±0.5 & 46.8±2.7 & 31.2±4.4 & 14.6±1.7 &  63.1±0.8 & 3.2±1.2 & 0.8±0.3 & 0.2±0.1  & 59.8±0.2 & 3.1±0.0 & 3.1±0.0 & \underline{3.1±0.0} \\
    &ContextPred & {ICLR'20}& 77.0±0.5 & 55.1±1.3 & 34.2±4.6 & 14.6±1.7  & 63.0±0.6 & 3.3±1.3 & 0.4±0.4 & 0.0±0.0 & 60.0±0.2 & 7.5±1.7 & 3.1±0.0 & \underline{3.1±0.0}\\
    &EdgePred &{ICLR'20}& 75.6±0.5 & 54.2±4.0 & 34.6±7.2 & 12.2±2.4  & 63.5±1.1 & 4.8±3.0 & 0.8±0.4 & 0.1±0.1  & 59.9±0.2 & 3.1±0.0 & 3.1±0.0 & \underline{3.1±0.0} \\
    &GraphCL& {NeurIPS'20} & 75.6±1.6 & 46.8±7.6& 32.2±6.8 &18.0±3.7& 52.2±0.2 & 0.5±0.3& 0.1±0.1 & 0.0±0.1 & 67.2±0.5 & 15.6±3.1 & 3.1±0.0 & \underline{3.1±0.0} \\
    &GROVER&{NeurIPS'20} & 73.3±1.4 & 38.5±2.0& 22.4±3.6 & 14.6±2.4 & 53.1±0.4 & 0.5±0.1& 0.1±0.1 & 0.0±0.1 & 66.2±0.1 & 15.6±0.0 & 3.8±1.4 & \underline{3.1±0.0} \\
    &JOAO&{ICML'21}  & 75.1±1.0 & 47.8±5.1 & 33.7±2.0 & 19.0±3.2& 52.3±0.2 & 0.4±0.1& 0.1±0.1 & 0.0±0.1  & 67.3±0.4 & 12.5±0.0 & 3.8±1.4 & \underline{3.1±0.0} \\
    &MGSSL &{NeurIPS'21}& 75.1±1.1 & 39.0±4.6 & 29.3±3.0 & 10.3±3.2 & 64.2±0.2 & 4.0±0.4& 1.7±0.3 &0.9±0.2& 66.9±0.5 & 13.8±2.8 & 3.1±0.0 & \underline{3.1±0.0} \\
    &GraphLoG & {ICML'21}    & 73.5±0.7 & 41.9±2.0  & 29.3±3.4 & 15.6±2.8 & 58.6±0.4 & 2.5±0.3 & 1.2±0.1&0.6±0.1& 62.9±0.4 & 4.4±1.7 & 0.0±0.0 & 0.0±0.0 \\
    &GraphMAE&{KDD'22} & 74.7±0.1 & 33.2±1.3& 27.8±1.3 & 12.2±1.7 & 53.3±0.1 & 0.6±0.1& 0.2±0.1 & 0.1±0.1 & 66.8±0.3 & 14.4±1.7 & 3.1±0.0 & \underline{3.1±0.0} \\
    &DSLA& {NeurIPS'22}  & 69.3±1.0 & 23.9±4.7 & 14.6±5.5 & 6.8±1.1 & 57.8±0.5 & 0.7±0.1 & 0.2±0.1 & 0.1±0.1 &  63.3±0.3 & 6.3±0.0 & 3.1±0.0 & \underline{3.1±0.0} \\
    \midrule
    % ---- Multi-Modal ----
    \multirow{8}{*}{\rotatebox{90}{\textbf{Multi-Modal}}}
    &Roberta-102M&- & 74.7±1.9 & 46.3±3.4 & 35.1±4.4 & 22.9±1.3 & 64.2±0.8 & 3.1±1.8 & 0.9±0.6 & 0.2±0.2 & 59.8±0.7 & 5.0±1.7 & 3.1±0.0 & \underline{3.1±0.0} \\
    &GPT2-87M &-& 71.0±3.4 & 31.2±11.2 & 20.0±9.4 & 7.3±6.9 & 61.5±1.1 & 2.4±0.6 & 0.8±0.3 & 0.3±0.1 & 60.6±0.3 & 7.5±1.7 & 1.9±1.7 & 1.9±1.7 \\
    &MolT5&{EMNLP'22} & 69.9±0.8 & 32.2±2.0 & 21.0±4.1 & 8.8±1.3 & 64.7±0.9 & 3.6±1.1 & 1.1±0.3 & 0.4±0.2 & 55.1±0.9 & 3.1±0.0 & 3.1±0.0 & 1.3±1.7 \\
    &UniMol&{ICLR'23} & 76.8±0.4 & 46.8±2.0  &33.7±1.1&24.9±2.0&64.6±0.2  &4.8±1.0 & 1.2±0.1&0.6±0.1& 65.4±0.1 & 7.5±1.7 & 3.1±0.0 &\underline{3.1±0.0} \\
    &CLOOME&{NC'23}  & 66.7±1.8 & 26.8±4.6 & 16.1±3.7 & 10.7±5.1  & 54.2±1.0 & 0.9±0.2 & 0.3±0.1 & 0.1±0.1 & 61.7±0.4 & 3.1±0.0 & 3.1±0.0 & 0.0±0.0 \\
    &InfoCORE (GE)&{ICLR'24} & 79.3±0.9 & 62.4±2.8& 46.3±3.0 & 30.3±2.2 & 65.3±0.2 & 5.4±1.7 & 1.2±0.4 & 0.3±0.1 & 60.2±0.2 & 3.1±0.0 & 0.0±0.0 & 0.0±0.0 \\
    &InfoCORE (CP) & ICLR'24& 73.8±2.0 & 37.6±9.2& 26.3±4.7 & 10.7±4.1 & 62.4±0.4 & 1.3±0.5 & 0.3±0.3 & 0.0±0.0 & 61.1±0.2 & 6.3±0.0 & 3.1±0.0 & 0.0±0.0 \\
    &InfoAlign&{ICLR'25} &  \underline{81.3±0.6} &  \underline{66.3±2.7} & \underline{49.3±2.7} & \underline{35.1±3.7} & \underline{66.4±1.1} & \underline{6.6±1.6}& \underline{2.1±0.5} & 0.7±0.3 & \underline{70.0±0.1} & \underline{18.8±2.2} & 3.1±0.0 & \underline{3.1±0.0} \\
    
    &Ours&-&{\textbf{84.7±0.2}} & {\textbf{73.2±0.0}} & {\textbf{66.8±1.3}} & {\textbf{45.9±2.0}} & {\textbf{69.3±0.3}} & {\textbf{16.1±0.6}}& {\textbf{7.5±0.5}} & {\textbf{3.0±0.5}} & {\textbf{71.4±0.2}} & {\textbf{24.4±1.4}} & {\textbf{6.2±0.0}}  & {\textbf{4.4±1.5}}  \\
    \bottomrule
  \end{tabular}
    \captionof{table}{We report classification performance on ChEMBL, ToxCast, and Broad, measured by average AUC\% (Avg.) and the percentage of tasks that achieve AUC\% thresholds of 80\%, 85\%, and 90\%. The \textbf{best} and \underline{second-best} average scores are highlighted.}
\label{tab2}%
\end{table*}

\begin{table*}[htbp]\small
  \centering
  \label{tab_che-tox}
  \renewcommand{\arraystretch}{0.9}
  \renewcommand{\tabcolsep}{1mm}
    \begin{tabular}{ccc|cc|ccccc|c}
    \toprule
        \multicolumn{3}{c|}{\textbf{Data Modality Type}} &
    \multicolumn{2}{c|}{\textbf{Dataset}} & \textbf{BBBP} & \textbf{BACE} & \textbf{ClinTox} & \textbf{SIDER} & \textbf{HIV} & \multirow{1.5}[2]{*}{\textbf{Avg.}} \\
    \multicolumn{3}{c|}{}&\multicolumn{2}{c|}{( \textbf{Molecule} /  \textbf{Task})} & \textbf{(2039/1)} & \textbf{(1513/1)} & \textbf{(1478/2)} & \textbf{(1427/27)} & \textbf{(41127/1)} &  \\
    \midrule
    \textbf{Single} &\textbf{Internal} &\textbf{External}&\textbf{Method} & \textbf{Venue} & \boldmath{}\textbf{\textbf{AUC\% $\uparrow$}}\unboldmath{} & \boldmath{}\textbf{\textbf{AUC\% $\uparrow$}}\unboldmath{} & \boldmath{}\textbf{\textbf{AUC\% $\uparrow$}}\unboldmath{} & \boldmath{}\textbf{\textbf{AUC\% $\uparrow$}}\unboldmath{} & \boldmath{}\textbf{\textbf{AUC\% $\uparrow$}}\unboldmath{} & \boldmath{}\textbf{\textbf{AUC\% $\uparrow$}}\unboldmath{} \\
    \midrule
    \ding{51}&\ding{55}&\ding{55}&AttentiveFP & JMC‘19 & 64.3±1.8 & 78.4±0.1 & 84.7±0.3 & 60.6±3.2 & 75.7±1.4 & 72.7±1.4 \\
    \ding{51}&\ding{55}&\ding{55}&PretrainGNN & ICLR’20 & 68.7±1.3 & 84.5±0.7 & 72.6±1.5 & 62.7±0.8 & 79.9±0.7 & 73.7±1.0 \\
    \ding{51}&\ding{55}&\ding{55}&GROVER & NeurIPS‘20 & 70.0±0.1 & 82.6±0.7 & 81.2±3.0 & 64.8±0.6 & 62.5±0.9 & 72.2±1.1 \\
    \ding{51}&\ding{55}&\ding{55}&MolCLR & NMI'22 & 72.2±2.1 & 82.4±0.9 & 91.2±3.5 & 58.9±1.4 & 78.1±0.5 & 76.6±1.7 \\
    \midrule
    \ding{55}&\ding{51}&\ding{55}&GEM   & NMI'22 & 72.4±0.4 & 85.6±1.1 & 90.1±1.3 & 67.2±0.4 & 80.6±0.9 & 79.2±0.8 \\
    \ding{55}&\ding{51}&\ding{55}&GraphMVP & ICLR'22 & 72.4±1.6 & 81.2±0.9 & 79.1±2.8 & 63.9±1.2 & 77.0±1.2 & 74.7±1.5 \\
   \ding{55}&\ding{51}&\ding{55}&UniMol & ICLR'23 & 72.9±0.6 & 85.7±0.2 & 91.9±1.8  & 65.9±1.3  & 80.8±0.3  & 79.4±0.8 \\
    % \ding{55}&\ding{51}&\ding{55}&Holi-Mol & TMLR'24 & 71.4±0.4  & 82.3±1.6  & 95.2±1.0 & 61.0±0.6  & 76.3±0.4 & 77.2±0.8 \\
    \ding{55}&\ding{51}&\ding{55}&MOLEBLEND &ICLR'24 & 73.0±0.8 & 83.7±1.4 & 87.6±0.7 & 64.9±0.3 & 79.0±0.8 & 77.6±0.8 \\
   \ding{55}&\ding{51}&\ding{55}&MOL-Mamba & AAAI'25 & \underline{75.0±0.2} & 86.4±0.4 & 92.7±1.1 & \underline{68.3±0.9} & \underline{81.6±0.5} & \underline{80.8±0.6} \\
    \midrule
    \ding{55}&\ding{51}&\ding{51}&InfoCORE(CP) & ICLR'24 & 74.0±0.8 & 85.0±0.2 & 89.3±0.5 & 65.2±0.1 & 78.7±0.1  & 78.4±0.3 \\
    \ding{55}&\ding{51}&\ding{51}&InfoCORE(GE) & ICLR'24 & 73.5±0.3 & \underline{86.6±0.3} & 91.9±1.9 & 64.8±0.6 & 78.5±0.2 & 79.1±0.7 \\
    \ding{55}&\ding{51}&\ding{51}&Atomas & ICLR'25 & 73.7±1.6 & 83.1±1.7 & \underline{93.1±0.5} & 64.4±1.9 & 80.5±0.4 & 79.0±1.2 \\
    \ding{55}&\ding{55}&\ding{51}&InfoAlign  & ICLR'25  & 73.5±0.6 & 83.2±0.2 & 89.4±0.7 & 64.1±0.3 & 81.5±0.2 & 78.3±0.4 \\
     
    \ding{55}&\ding{51}&\ding{51}&Ours  & -  & {\textbf{75.2±0.1}} & {\textbf{88.7±0.3}} & {\textbf{94.4±0.2}} & {\textbf{70.3±0.4}} & {\textbf{82.5±0.1}} & {\textbf{82.2±0.2}} \\
    \bottomrule
    \end{tabular}%
    \captionof{table}{Classification performance on the BBBP, BACE, ClinTox, SIDER, and HIV datasets. The \textbf{best} and \underline{second-best} average scores are highlighted.}
    \label{tab3}%
\end{table*}%

\section{Downstream Fine-Tuning Setup} \label{sec:finetune_setup}

In molecular property prediction tasks, we adopt a lightweight fine-tuning strategy to preserve the multi-modal representations learned during pretraining while adapting to specific downstream datasets.

Specifically, given a molecule $v_i$, we first extract its projected multi-modal representations:

\begin{equation}
\begin{aligned}
\mathbf{p}^{\mathrm{1D}} &= f^{\mathrm{1D}}(\mathbf{x}^{1D}_i),\\[0.8ex]
\mathbf{p}^{\mathrm{2D}} &= f^{\mathrm{2D}}\big(\underbrace{g_{\mathrm{GNN}}(G_i)}_{\text{pre-trained}}\big),\\[0.8ex]
\mathbf{p}^{\mathrm{3D}} &= f^{\mathrm{3D}}(\mathbf{x}^{3D}_i),
\end{aligned}
\end{equation}
where $f^{\mathrm{1D}}, f^{\mathrm{2D}}, f^{\mathrm{3D}}$ denotes the pretrained 1D fingerprint, molecular graph, and 3D conformation modality projectors, respectively. $g_{\mathrm{GNN}}$ is a GNN-based molecular structure encoder used to extract topological information from the molecular graph $G_i$, producing structural features $\mathbf{x}^{2D}_i$. In our implementation, $g_{\mathrm{GNN}}$ adopts the Virtual GNN architecture~\cite{hwang2022analysis} pretrained on large-scale molecular datasets.  

These modality features are concatenated as:  

\begin{equation}
\mathbf{z}_i = \mathrm{Concat}\left(\mathbf{x}^{1D}_i,\, \mathbf{p}^{\mathrm{1D}},\, \mathbf{p}^{\mathrm{2D}},\, \mathbf{p}^{\mathrm{3D}}\right),
\end{equation}

Under the condition of freezing all backbone parameters ($g_{\mathrm{GNN}}, f_{\mathrm{1D}}, f_{\mathrm{2D}}, f_{\mathrm{3D}}$), we train a lightweight task-specific decoder $g_{\theta}(\cdot)$, implemented as a two-layer MLP:  

\begin{equation}
\hat{y}_i = g_{\theta}(\mathbf{z}_i),
\end{equation}

\section{Experiment Details} \label{sec:exp_details}

\subsection{Dataset}

We pretrain our model on a large-scale multi-modal molecular dataset constructed from diverse public resources, including the DrugBank molecular library \cite{wishart2018drugbank}, Cell Painting images \cite{bray2016cell}, the JUMP-CP multi-omics platform \cite{chandrasekaran2023jump}, and L1000 gene expression profiles \cite{wang2016drug, subramanian2017next}. This dataset is represented as a heterogeneous molecule–gene–cell graph, covering 129,592 molecules with paired biological responses \cite{liu2025learning}. The molecular modalities include \textbf{1D fingerprints} (concatenation of ECFP and MACCS descriptors), \textbf{2D graph structures} (atom-level topological features extracted using Virtual GNN~\cite{hwang2022analysis}), and \textbf{3D spatial conformations} (pretrained representations from UniMol~\cite{zhou2023unimol}). These structural modalities are fully available for all molecules. In contrast, the external biological modalities are highly incomplete and asymmetric (Table~\ref{tab:modality_missing}). Specifically, \textbf{cellular modalities} include CP-JUMP and CP-Bray, two Cell Painting datasets capturing compound-induced morphological profiles \cite{bray2016cell, chandrasekaran2023jump}. CP-JUMP covers most molecules (missing rate 12.5\%), whereas CP-Bray is highly sparse (91.4\% missing). \textbf{Gene perturbation modalities}, G-CRISPR and G-ORF, provide cellular responses to CRISPR knockouts and ORF overexpression, respectively, but are largely incomplete (missing rates $>$99\%). Finally, the \textbf{gene expression modality} from L1000 profiles captures transcriptomic responses of 978 landmark genes but is missing for 94.3\% of molecules.

We evaluate our framework on \textbf{nine benchmark datasets}, covering eight classification tasks and one regression task. Together, they comprise \textbf{728 prediction tasks} spanning toxicity, pharmacokinetics, target specificity, and more. Data statistics are summarized in Table~\ref{datasets}. A brief summary of each dataset is as follows:

\paragraph{Classification Tasks}
\textbf{ChEMBL}~\cite{gaulton2012chembl}: Contains 41 binary classification tasks on binding affinity and activity annotations derived from the ChEMBL database.
\textbf{ToxCast}~\cite{richard2016toxcast}: Comprises 617 binary classification tasks for toxicity profiling under the “Toxicology in the 21st Century” initiative.
\textbf{Broad}~\cite{moshkov2023predicting}: Focuses on predicting compound mechanisms of action (MoA) based on cellular response data.
\textbf{BACE}~\cite{hu2020open}: Predicts inhibitory activity against $\beta$-secretase (BACE), a target relevant for Alzheimer’s disease drug discovery.
\textbf{BBBP}~\cite{hu2020open}: Predicts blood–brain barrier permeability of small molecules to assess CNS accessibility.
\textbf{ClinTox}~\cite{hu2020open}: Distinguishes FDA-approved drugs from those withdrawn due to toxicity, aiding clinical safety evaluation.
\textbf{SIDER}~\cite{hu2020open}: A multi-label classification task based on adverse drug reactions curated in the SIDER database.
\textbf{HIV}~\cite{hu2020open}: Predicts HIV replication inhibition activity of candidate compounds.

\paragraph{Regression Tasks}
\textbf{Biogen}~\cite{fang2023prospective}: Consists of six ADME (Absorption, Distribution, Metabolism, Excretion) properties, including HLM/RLM stability, MDR1-MDCK efflux ratio (ER), solubility at pH 6.8, and plasma protein binding (rPPB) in human/mouse.

\subsection{Baselines for Comparison}

To validate the effectiveness of our method, we compare against \textbf{over 20 baseline models} grouped into four categories:

\paragraph{Fingerprint Models.}  
\textbf{Molecular fingerprint}~\cite{rogers2010extended}: Classical cheminformatics descriptors encoding molecular substructures.\\
\textbf{AttentiveFP}~\cite{xiong2019pushing}: A fingerprint-based model for learning task-specific molecular representations.

\paragraph{Single-Modality Methods.}  
\textbf{PretrainGNN}~\cite{hu2020strategies} combines multiple pretraining strategies—including \textbf{AttrMask}, \textbf{ContextPred}, and \textbf{EdgePred}—to capture complementary local and global graph semantics, thereby enhancing the transferability of molecular graph representations.\\
\textbf{GraphCL}~\cite{you2020graph}: Contrastive learning on molecular graphs with graph-level augmentations.\\
\textbf{GROVER}~\cite{rong2020self}: Combines GNNs and Transformers for context-aware molecular representations.\\
\textbf{JOAO}~\cite{you2021graph}: Jointly optimizes graph augmentations in a contrastive learning framework.\\
\textbf{MGSSL}~\cite{zhang2021motif}: Exploits motif-based contrastive pretraining for molecular graphs.\\
\textbf{GraphLoG}~\cite{xu2021self}: Learns graph-level representations with subgraph-based contrastive learning.\\
\textbf{GraphMAE}~\cite{hou2022graphmae}: A masked autoencoder for molecular graph pretraining.\\
\textbf{DSLA}~\cite{kim2022graph}: Dual-stage pretraining for improved structure learning on graphs.\\
\textbf{MolCLR}~\cite{wang2022molecular}: Uses chemically aware perturbations for contrastive learning on molecular graphs.

\paragraph{Molecular Multimodal Methods.}  
\textbf{GEM}~\cite{fang2022geometry}: Learns joint molecular representations from graphs and 3D geometries.\\
\textbf{GraphMVP}~\cite{liu2022pre}: Cross-modal contrastive learning between 2D graphs and 3D conformations.\\
\textbf{UniMol}~\cite{zhou2023unimol}: SE(3)-equivariant Transformer for unified modeling of molecular conformations.\\
\textbf{MolT5}~\cite{edwards2022translation}: A Transformer language model pretrained on SMILES for molecular tasks.\\
\textbf{RoBERTa-102M/GPT2-87M}~\cite{datamol2024}: Large-scale language models fine-tuned for molecular representations.\\
\textbf{MOLEBLEND}~\cite{yu2024multimodal}: Blends multiple molecular modalities for fine-grained feature alignment.\\
\textbf{MOL-Mamba}~\cite{hu2025mol}: Combines electronic semantics with graph state-space models.

\paragraph{Biological Multimodal Methods.}  
\textbf{CLOOME}~\cite{sanchez2023cloome}: Aligns molecular structures with cell painting images using contrastive learning.\\
\textbf{InfoCORE}~\cite{wang2023removing}: Mitigates batch effects for robust molecule–cell alignment.\\
\textbf{Atomas}~\cite{zhang2025atomas}: Introduces hierarchical alignment between atomic, fragment, and molecular levels with textual data.\\
\textbf{InfoAlign}~\cite{liu2025learning}: Jointly models molecular, cellular, and transcriptomic features for mechanism-of-action prediction, while reducing redundancy through mutual information maximization.

\subsection{Implementation Details}
\paragraph{Pretraining}
We perform pretraining on an NVIDIA RTX 3090 GPU. The model is trained for 50 epochs using the Adam optimizer~\cite{kingma2014adam} with a learning rate of $1 \times 10^{-4}$ and weight decay of $1 \times 10^{-8}$. A cosine learning rate scheduler with linear warm-up~\cite{loshchilov2016sgdr} is applied. The batch size is set to 3072. The context graph follows a random walk of length $L=4$, and feature propagation is trained for $T=5$ iterations, using $K=10$ nearest neighbors. The hyperparameters $\eta$, $H$, $\lambda_1$, and $\lambda_2$ are adjusted via hyperparameter search, as detailed in Figure~3.

The model architecture includes the following components:
\begin{itemize}
    \item \textbf{GNN Backbone}: We adopt a 5-layer Graph Isomorphism Network (GIN) \cite{xu2018powerful} as the structural encoder for molecular graphs. To enhance message passing and facilitate global information flow, we incorporate a virtual node mechanism \cite{hwang2022analysis} that adds a dedicated embedding per graph. In each GIN layer, the following steps are performed: (1) virtual node features are added to the node embeddings to inject graph-level context; (2) message passing is performed via edge-aware GIN convolution; (3) layer normalization (BatchNorm or GraphNorm) is applied to stabilize training; (4) activation and dropout are applied, except in the final layer; and (5) residual connections are used to enhance gradient flow. 
    \item \textbf{Modality Projectors}: Each modality (1D, 2D, 3D, and external biological) is projected into a shared latent space using a lightweight MLP with a hidden dimension of 300, consisting of a LayerNorm, a linear layer, and a SiLU activation.
    \item \textbf{Modality Decoders}: Each modality uses a two-layer MLP decoder with a hidden dimension of 1200, consisting of LayerNorm, GELU activation, dropout, and linear transformations, to reconstruct original input features from the projected embeddings.
\end{itemize}

\paragraph{Fine-Tuning}
For downstream fine-tuning tasks, $\eta = 1$ and $H = 6$ are fixed, while $\lambda_1$ and $\lambda_2$ are adapted for the specific downstream molecular baseline datasets. The GNN backbone and the 1D, 2D, and 3D modality projectors are all frozen during downstream fine-tuning. In fine-tuning, hyperparameters such as learning rate, dropout, and hidden layer size play a crucial role in balancing overfitting and generalization.

For four datasets—BACE, ClinTox, SIDER, and HIV—we introduce a \textbf{structure-aware ensemble strategy} to incorporate prior knowledge from traditional cheminformatics features. Specifically, we train a lightweight machine learning model $M_{\mathrm{RF}}(\cdot)$, such as random forests, on the traditional 1D fingerprints $\mathbf{x}^{(1D)}$ to obtain prior predictions:

\begin{equation}
\hat{y}_i^{\mathrm{RF}} = M_{\mathrm{RF}}(\mathbf{x}^{(1D)}_i).
\end{equation}

The final prediction is obtained by combining the CHMR output with the prior prediction using a weighted average. The ensemble weight $\gamma$ controls the balance between the two components:

\begin{equation}
\hat{y}_i^{\mathrm{final}} = \gamma \hat{y}_i^{\mathrm{CHMR}} + (1-\gamma)\hat{y}_i^{\mathrm{RF}},
\end{equation}
where $\hat{y}_i^{\mathrm{CHMR}}$ denotes the prediction from the CHMR framework, and $\gamma\in [0,1]$ is a balancing hyperparameter applied to selected datasets for structure-aware integration.

Table~\ref{tab:hyperparam_settings} summarizes the hyperparameter settings for all nine benchmark datasets, detailing the settings for both pretraining and fine-tuning.

\begin{figure*}[htpb]
    \centering
    \includegraphics[width=1\textwidth]{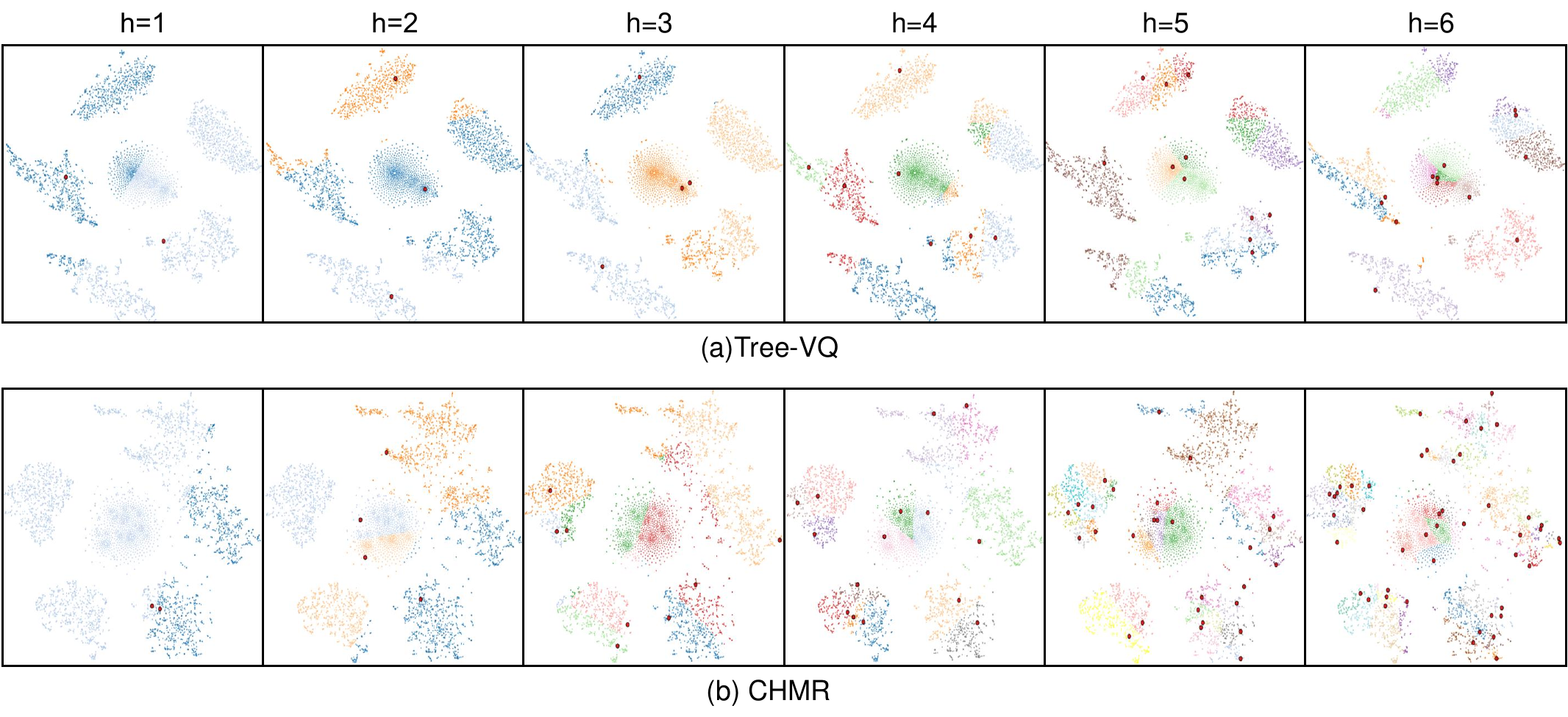}
    \caption{Hierarchical Center Embedding Visualization.
    Comparison of Tree-VQ (top row) and CHMR (bottom row) across tree levels \( h = 1 \) to \( h = 6 \) (left to right). CHMR shows denser and more discriminative clusters with deeper and more balanced tree utilization, reflecting improved modeling of hierarchical semantics.}
    \label{fig:appendix_treevq_chmr}
\end{figure*}

\subsection{Extended Results on Molecular Property Prediction}

To further evaluate the model's performance in high-reliability scenarios, Table \ref{tab2} reports the proportion of tasks exceeding high AUC thresholds (80\%, 85\%, 90\%) on three classification datasets. This reflects the stability and broad applicability of the model under higher precision requirements. The results show that our method significantly outperforms in all threshold ranges, indicating stronger robustness and generalization.

Table \ref{tab3} further extends to a broader range of molecular property prediction, including BBBP, BACE, ClinTox, SIDER, and HIV. We categorize the comparison methods based on the types of modalities used: single-modality methods, internal structure multimodal methods, and structure + phenotype external multimodal methods. Our method achieves the best performance across all datasets, with an average AUC of 82.2\%, outperforming the second-best methods, InfoAlign (79.1\%) and MOL-Mamba (80.8\%). Notably, on multi-label tasks such as ClinTox and SIDER, our method improves by approximately 2.0–3.0\%.

In summary, our method demonstrates leading performance in molecular property and function prediction tasks. It achieves optimal results in both classification and regression tasks, indicating strong modeling capabilities; maintains a higher task coverage under higher AUC thresholds, reflecting robust performance; and shows excellent generalization in more datasets and complex baseline scenarios.

\subsection{Additional Visualizations of Tree-VQ and CHMR Representations}
Figure~\ref{fig:appendix_treevq_chmr} provides a full-layer visualization of the hierarchical codebook activations learned by Tree-VQ (top row) and CHMR (bottom row). Each column corresponds to a specific tree level $h \in \{1, 2, \dots, 6\}$, ranging from the coarsest level (left, $h=1$) to the most fine-grained level (right, $h=6$). Dots represent sample projections, and red bolded nodes indicate the quantization centers used at that level.

For Tree-VQ, we observe under-utilization of deep-layer codes and excessive concentration around a few centers, suggesting limited semantic differentiation. In contrast, CHMR demonstrates more balanced code usage across all levels and produces clusters that become progressively refined from $h=1$ to $h=6$, indicating its stronger capacity to model hierarchical semantic structure across modalities.

\end{document}